\documentclass[runningheads]{llncs}

\usepackage{eccv}

\usepackage{eccvabbrv}

\usepackage{graphicx}
\usepackage{booktabs}
\usepackage{amsmath}
\usepackage{amssymb}
\usepackage{subcaption}
\usepackage{xspace}

\usepackage[accsupp]{axessibility}

\usepackage[breaklinks,colorlinks,citecolor=eccvblue]{hyperref}

\newcommand{\mine}{{\small \sf \mbox{\emph{CORE}}}\xspace}

\begin{document}

\title{CORE: Robust Out-of-Distribution Detection via Confidence and Orthogonal Residual Scoring}

\titlerunning{CORE}

\author{Jin Mo Yang\inst{1} \and Hyung-Sin Kim\inst{1} \and Saewoong Bahk\inst{1}}
\authorrunning{J.M. Yang et al.}
\institute{Seoul National University, Seoul, South Korea\\
\email{\{jmyang, hyungkim, sbahk\}@snu.ac.kr}}

\maketitle

\begin{abstract}
\label{sec:abstract}
Out-of-distribution (OOD) detection is essential for deploying deep learning models reliably, yet no single method performs consistently across architectures and datasets---a scorer that leads on one benchmark often falters on another. We attribute this inconsistency to a shared structural limitation: logit-based methods see only the classifier's confidence signal, while feature-based methods attempt to measure membership in the training distribution but do so in the full feature space where confidence and membership are entangled, inheriting architecture-sensitive failure modes. We observe that penultimate features naturally decompose into two orthogonal subspaces: a classifier-aligned component encoding confidence, and a residual the classifier discards. We discover that this residual carries a class-specific directional signature for in-distribution data---a membership signal invisible to logit-based methods and entangled with noise in feature-based methods. We propose \mine (COnfidence + REsidual), which disentangles the two signals by scoring each subspace independently and combines them via normalized summation. Because the two signals are orthogonal by construction, their failure modes are approximately independent, producing robust detection where either view alone is unreliable. \mine achieves competitive or state-of-the-art performance across five architectures and five benchmark configurations, ranking first in three of five settings and achieving the highest grand average AUROC with negligible computational overhead.

\end{abstract}

\section{Introduction}\label{sec:intro}

\begin{figure*}[t]
    \centering
    \begin{subfigure}[t]{0.62\linewidth}
        \centering
        \includegraphics[width=\linewidth]{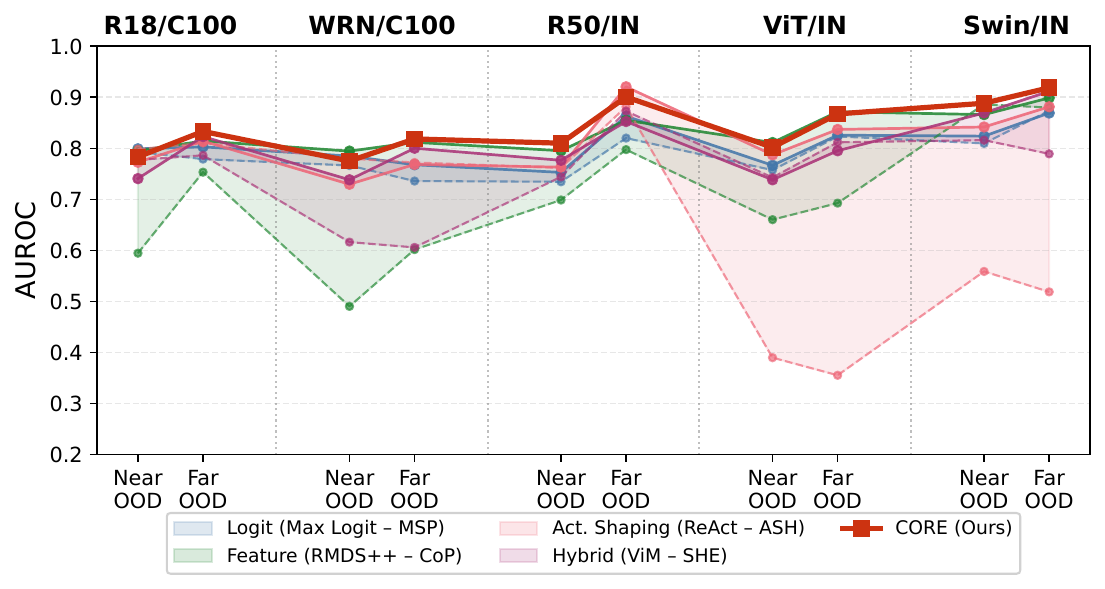}
        \caption{Near-OOD and far-OOD AUROC across five model$\times$ID settings. Shaded bands span the best-to-worst scorer within each category; dashed lines mark the category worst.}
        \label{fig:intro-nearfar}
    \end{subfigure}
    \hfill
    \begin{subfigure}[t]{0.36\linewidth}
        \centering
        \includegraphics[width=\linewidth]{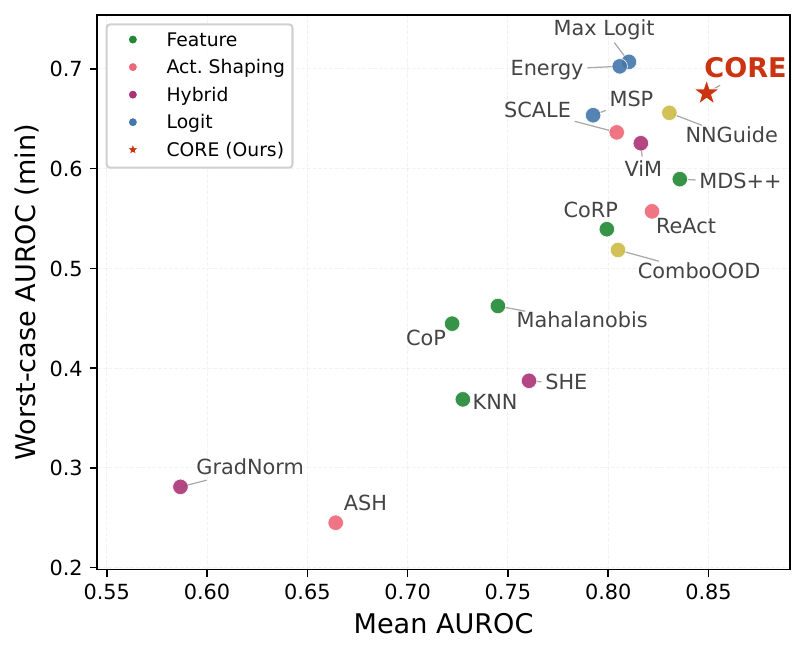}
        \caption{Mean vs.\ worst-case AUROC. The top-right corner is ideal.}
        \label{fig:intro-scatter}
    \end{subfigure}
    \caption{\textbf{Fragility of post-hoc OOD scorers across architectures and datasets.} Catastrophic failures occur systematically for activation-shaping and feature-based methods on specific architectures (e.g., ViT/IN, Swin/IN). \mine (red) is consistently the most robust scorer across all settings.}
    \label{fig:intro}
\end{figure*}

Out-of-distribution (OOD) detection is essential for deploying deep learning models reliably in open-world environments~\cite{yang2024generalizedoutofdistributiondetectionsurvey}.
Among the many proposed approaches, \emph{post-hoc} scoring methods are particularly attractive: they require no retraining, apply to any pretrained classifier, and operate at negligible cost during inference~\cite{hendrycks2016baseline,liu2020energy}.
Yet a persistent and under-discussed problem remains---\textbf{no single post-hoc method is reliably the best across architectures and datasets.}
Energy~\cite{liu2020energy} dominates on some benchmarks, KNN~\cite{sun2022out} on others, ViM~\cite{wang2022vim} on yet others.
A method that ranks first in one setting can drop to mediocre performance in the next.
For practitioners, this inconsistency is a serious obstacle: without knowing which OOD inputs will be encountered, there is no principled way to choose a scorer.

Figure~\ref{fig:intro} shows the comprehensive picture.
We evaluate the best and worst scorers from each of five baseline categories---logit-based, feature-based, activation shaping, hybrid, and score combination---across five model$\times$ID configurations with both near-OOD and far-OOD benchmarks (Fig.~\ref{fig:intro-nearfar}).
Shaded bands show the full range within each category.
Catastrophic failures are systematic, not isolated: activation-shaping methods collapse on ViT/ImageNet and Swin/ImageNet, while feature-based methods exhibit wide variance across all settings.
Fig.~\ref{fig:intro-scatter} summarizes each scorer by its mean and worst-case AUROC: \mine (red star) sits in the upper-right corner, achieving the highest mean AUROC while maintaining the best worst case.
The question is: \emph{why does every category have a blind spot, and can we close them?}

Existing methods each rely on a different criterion---and each breaks down in certain settings (Fig.~\ref{fig:root-cause}).
Logit-based methods~\cite{hendrycks2016baseline,liu2020energy} measure \emph{confidence} but are blind to OOD samples that produce confident predictions (Fig.~\ref{fig:root-cause}a).
Feature-based methods~\cite{lee2018mahalanobis,sun2022out} measure \emph{membership} via distances in feature space, but operate on the full $z$ where confidence and membership are entangled, inheriting architecture-sensitive pathologies such as hubness (Fig.~\ref{fig:root-cause}b).
Activation shaping methods~\cite{sun2021react,djurisic2023ash} denoise features before recomputing logits, but their heuristics vary across architectures (Fig.~\ref{fig:root-cause}c).

The common thread is that the feature $z$ carries two distinct signals---confidence and membership---and no existing method cleanly separates them.
We formalize this observation.
A classifier's final linear layer computes logits $\ell = Wz$, where $W \in \mathbb{R}^{C \times d}$ and $z \in \mathbb{R}^d$ is the penultimate feature.
This operation is a projection: it uses only the component of $z$ aligned with class weight directions.
We can decompose any feature $z$ with respect to its predicted class $\hat{y}$ into two orthogonal components:
\begin{figure*}[t]
    \centering
    \includegraphics[width=\linewidth]{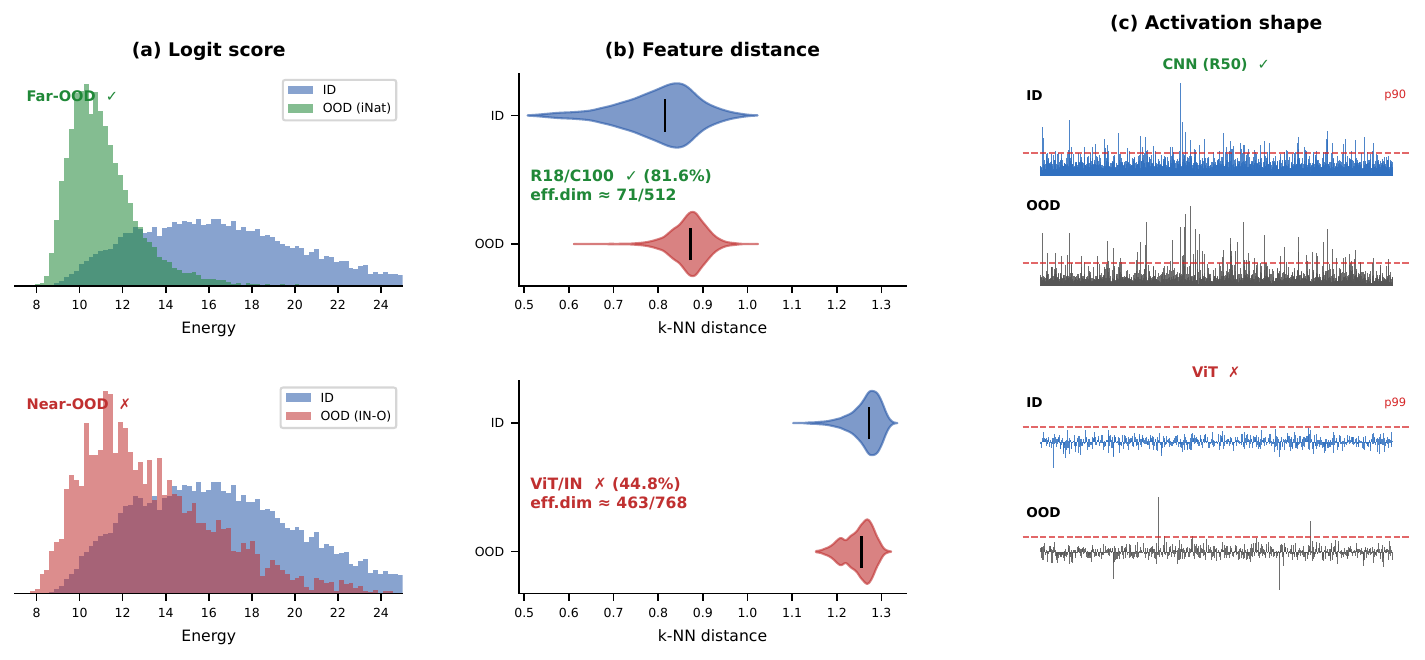}
    \caption{Each column represents one scorer category: (a)~logit-based energy score (far-OOD separates, near-OOD overlaps), (b)~feature-based k-NN distance (R18/C100 separates, ViT/IN concentrates), (c)~per-unit penultimate activations (CNN shows sparse OOD spikes above the ReAct threshold; ViT activations are symmetric with no clear ID/OOD separation). Top rows: settings where the approach succeeds; bottom rows: where it fails.}
    \label{fig:root-cause}
\end{figure*}

\begin{equation}\label{eq:decomp_intro}
    z = z_\parallel + z_\perp, \quad
    z_\parallel = \frac{z \cdot w_{\hat{y}}}{\|w_{\hat{y}}\|^2}\,w_{\hat{y}}, \quad
    z_\perp = z - z_\parallel,
\end{equation}
where $w_{\hat{y}}$ is the classifier weight for the predicted class.
The parallel component $z_\parallel$ determines the logit magnitude---it encodes \emph{confidence}.
The orthogonal residual $z_\perp$ is everything the classifier discards.

Through this lens, the blind spots identified above become precise: logit-based methods see only $z_\parallel$; feature-based methods measure membership in the full $z$ where it is entangled with $z_\parallel$; activation shaping modifies $z$ to improve $z_\parallel$ but never accesses $z_\perp$.
No existing method scores $z_\perp$ independently.

\noindent\textbf{Key finding: structured residual.} 
We discover that $z_\perp$ carries a \emph{class-specific directional signature} for in-distribution data.
Within each predicted class $c$, ID samples' residuals $z_\perp$ consistently align with a characteristic mean direction $\mu_\perp^{(c)}$, estimated from as few as ${\sim}13$ training samples per class.
OOD samples' residuals do not exhibit this alignment.
A single cosine similarity $\cos(z_\perp,\, \mu_\perp^{(\hat{y})})$ is therefore a powerful \emph{membership} signal---one that is entirely invisible to logit-based methods and entangled with noise in feature-based methods.

\noindent\textbf{Why combining works: orthogonal decoupling.} 
Because $z_\parallel \perp z_\perp$ by construction, the confidence signal (Energy score, derived from $z_\parallel$) and the membership signal ($\cos(z_\perp, \mu_\perp^{(\hat{y})})$) have approximately decoupled failure modes.
Confidence fails when OOD inputs produce high logits; membership fails when $z_\perp$ accidentally aligns with the ID direction.
These events are approximately independent ($r{=}0.30$--$0.40$, \S\ref{sec:analysis}).
Summing the two $z$-score-normalized signals provides OR semantics: \emph{either} anomaly triggers detection.
This is why \mine is robust where methods that see only one signal or both entangled, are brittle.

Our contributions are threefold:
\begin{itemize}
    \item We identify orthogonal subspace decomposition of classifier features as a principled framework that explains \emph{why} existing OOD methods are inconsistent---logit-based methods see only confidence ($z_\parallel$), while feature-based methods measure membership entangled with confidence in the full $z$.

    \item We discover that the orthogonal residual $z_\perp$ carries a class-specific directional signature that cleanly separates ID from OOD, providing a membership signal invisible to logit-based methods and entangled with noise in feature-based methods.

    \item \mine combines confidence and residual membership via $z$-score summation with $\mathcal{O}(d)$ overhead, achieving the highest grand average AUROC (84.9\%) across five model$\times$dataset settings, ranking first in three of five, and never catastrophically failing where every baseline category does.
\end{itemize}

\section{Related Work}\label{sec:related}
We focus on post-hoc OOD detection methods that operate on pretrained classifiers without retraining.
We organize prior work by how each method accesses the feature vector $z$ and its derived quantities, and critique each category through the orthogonal decomposition $z = z_\parallel + z_\perp$ introduced in \S\ref{sec:intro}.

\subsection{Logit-Based Methods}
The simplest family of post-hoc scorers operates solely on the classifier output $\ell = Wz$.
MSP~\cite{hendrycks2016baseline} uses the maximum softmax probability, while Energy~\cite{liu2020energy} computes the log-sum-exp of logits.
MaxLogit~\cite{basart2022scaling} directly uses the largest logit value, and ODIN~\cite{liang2017enhancing} enhances MSP via temperature scaling and input perturbation.
These methods are simple, require no training data, and are the most architecture-consistent category in our benchmarks.
However, they share a fundamental blind spot: because $\ell = Wz$ captures only the classifier-aligned component $z_\parallel$, any OOD sample that produces confident logits---\eg, semantically near-OOD images---is invisible to these methods.

\subsection{Feature-Based Methods}
A second family computes scores from the penultimate feature $z$ using statistics estimated from ID training data.
Mahalanobis distance~\cite{lee2018mahalanobis} fits class-conditional Gaussians and measures distance to the nearest class mean; its variant MDS++~\cite{ren2021simple} applies L2-normalization to improve conditioning.
KNN~\cite{sun2022out} uses the $k$-th nearest neighbor distance in feature space, and CoP/CoRP~\cite{fang2024cop} measure cosine similarity between test features and class prototypes.
These methods attempt to measure membership---whether a test sample belongs to the training distribution---by computing distances or densities in feature space.
However, they operate on the full $z$ without separating the confidence signal ($z_\parallel$) from the membership signal ($z_\perp$); the two remain \emph{entangled}, and the full dimensionality of $z$ introduces architecture-sensitive failure modes: Mahalanobis distance suffers from covariance collapse in high dimensions~\cite{ren2021simple}, and KNN is susceptible to \emph{hubness}---a high-dimensional pathology where a few data points become nearest neighbors of disproportionately many others, causing OOD points to appear deceptively close to ID clusters.
Our benchmarks show that feature-based methods exhibit the largest CNN-to-transformer performance gap among all categories, confirming their sensitivity to feature geometry.

\subsection{Activation Shaping Methods}
A third family modifies feature activations before recomputing logits.
ReAct~\cite{sun2021react} clamps activations above a threshold, suppressing outlier magnitudes that OOD inputs tend to produce.
ASH~\cite{djurisic2023ash} zeroes out activations below a fixed percentile, and SCALE~\cite{xu2024scale} prunes and exponentially rescales activations.
These methods effectively denoise the feature vector to improve the logit-based signal.
However, shaping rules calibrated for one architecture can be destructive for another: ASH assumes sparse activations (as produced by ReLU networks), but transformer features are uniformly distributed due to LayerNorm, causing ASH to drop to near-chance AUROC on ViT/ImageNet.

\subsection{Hybrid Methods}
Several methods compute a single score that inseparably mixes feature-based and logit-based quantities.
ViM~\cite{wang2022vim} projects features onto the classifier's null space and combines the residual norm with an energy term; it comes closest to isolating membership by operating in the null space, but measures a \emph{global} residual norm rather than a \emph{class-conditional direction}, limiting its discriminative power.
SHE~\cite{zhang2023she} matches test features against memorized class-specific patterns via Hopfield energy.
GradNorm~\cite{huang2021importance} uses the gradient norm of the KL divergence between the softmax output and a uniform distribution.
These methods produce a single indivisible score from features and logits jointly, so their failure modes cannot be isolated or corrected independently.

\subsection{Score Combination Methods}
A distinct family explicitly fuses separately defined component scores.
NNGuide~\cite{park2023nnguide} multiplies the Energy score by a $k$-NN similarity weight, effectively using neighborhood density to gate classifier confidence.
ComboOOD~\cite{rajasekaran2024combood} sums Mahalanobis distance and $k$-NN scoring, fusing a parametric and a non-parametric distance measure.
These are the closest in spirit to \mine---they recognize that a single signal is insufficient and fuse complementary views.
However, their fusion is \emph{motivated but structurally fragile}: both methods extract their component signals from the same full feature $z$, so when the representation suffers from pathologies---covariance collapse in moderate dimensions or distance concentration (hubness) in high dimensions---both components degrade together (\S\ref{sec:analysis}).

\paragraph{Positioning.}
\mine differs from all five categories by decomposing $z$ into orthogonal subspaces $z_\parallel$ and $z_\perp$ and scoring each independently, yielding approximately independent failure modes by construction rather than by empirical coincidence.

\section{Method}\label{sec:method}

\begin{figure}[t]
    \centering
    \begin{subfigure}[b]{0.52\linewidth}
        \centering
        \includegraphics[width=\linewidth]{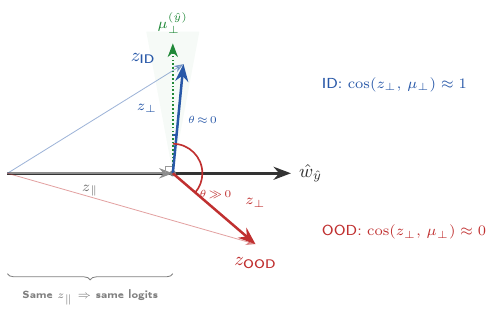}
        \caption{Both ID and OOD share the same $z_\parallel$ (identical logits), but their $z_\perp$ directions differ: ID aligns with $\mu_\perp^{(\hat{y})}$, OOD does not.}
        \label{fig:method-geom}
    \end{subfigure}
    \hfill
    \begin{subfigure}[b]{0.45\linewidth}
        \centering
        \includegraphics[width=\linewidth]{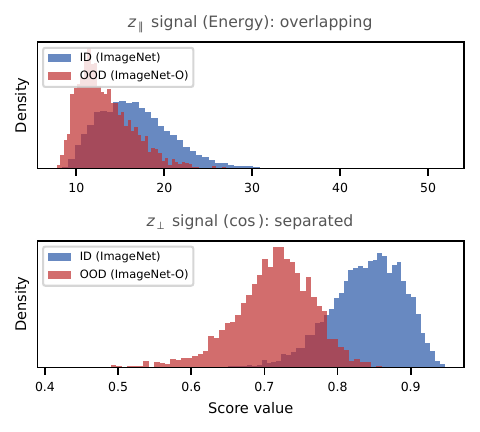}
        \caption{Score distributions (ResNet-50, ImageNet vs.\ ImageNet-O). Energy from $z_\parallel$ overlaps; cosine from $z_\perp$ separates.}
        \label{fig:method-hist}
    \end{subfigure}
    \caption{\mine decomposes features into classifier-aligned $z_\parallel$ and orthogonal residual $z_\perp$. The residual carries a membership signal that is more discriminative than confidence alone and invisible to logit-based methods.}
    \label{fig:method-overview}
\end{figure}

We present \mine, a scoring framework that decomposes penultimate features into two orthogonal subspaces and scores each independently.
Figure~\ref{fig:method-overview} provides an overview.

\subsection{Problem Setup}

Consider a pretrained classifier that maps an input image $x$ to a predicted class $\hat{y} \in \{1, \dots, C\}$.
The network consists of a feature extractor $\phi: \mathbb{R}^m \to \mathbb{R}^d$ producing a penultimate feature $z = \phi(x)$, and a linear classifier with weights $W \in \mathbb{R}^{C \times d}$ producing logits $\ell = Wz \in \mathbb{R}^C$.
The predicted class is $\hat{y} = \arg\max_c\, \ell_c$.

Post-hoc OOD detection defines a scoring function $S(x) \in \mathbb{R}$ such that higher values indicate in-distribution samples.
A sample is classified as OOD when $S(x) < \gamma$ for some threshold $\gamma$.
The goal is to design $S$ that is \emph{consistently} discriminative across diverse architectures and datasets.

\subsection{Orthogonal Feature Decomposition}

The key observation behind \mine is that the classifier uses only a one-dimensional projection of $z$ to compute the logit for any given class.
For the predicted class $\hat{y}$ with weight vector $w_{\hat{y}} \in \mathbb{R}^d$ (the $\hat{y}$-th row of $W$), the logit is $\ell_{\hat{y}} = w_{\hat{y}}^\top z$.
This means the classifier's decision depends only on the component of $z$ aligned with $w_{\hat{y}}$, discarding the $(d{-}1)$-dimensional orthogonal complement.

We formalize this by decomposing $z$ into two orthogonal components:
\begin{equation}\label{eq:decomp}
    z = z_\parallel + z_\perp, \quad
    z_\parallel = \frac{z^\top w_{\hat{y}}}{\|w_{\hat{y}}\|^2}\, w_{\hat{y}}, \quad
    z_\perp = z - z_\parallel.
\end{equation}
By construction, $z_\parallel^\top z_\perp = 0$.
The parallel component $z_\parallel$ encodes \emph{confidence}: it determines the logit magnitude for the predicted class and, through cross-talk with other class weights, influences the full logit vector.
The orthogonal residual $z_\perp$ encodes everything the classifier ignores---we show next that it carries a \emph{membership} signal.

\subsection{Confidence Score: Energy}\label{sec:confidence}

For the confidence component, we adopt the Energy score~\cite{liu2020energy}:
\begin{equation}\label{eq:energy}
    S_\text{conf}(x) = \log \sum_{c=1}^{C} \exp(\ell_c) = \log \sum_{c=1}^{C} \exp(w_c^\top z).
\end{equation}
Through the lens of our decomposition, Energy aggregates the projections of $z$ onto all class weight directions.
It is fundamentally a function of $z_\parallel$ across the $C$ class axes: it measures how strongly $z$ aligns with the classifier's learned directions overall.
ID samples that activate class neurons strongly receive high Energy; OOD samples that fall between class directions receive low Energy.
Energy is a well-established score~\cite{liu2020energy}; our contribution is to reinterpret it as a confidence signal operating in the $z_\parallel$ subspace and to complement it with an independent membership signal from $z_\perp$.

\subsection{Residual Membership Score}\label{sec:membership}

We now present the core novel component of \mine.
We discover that the orthogonal residual $z_\perp$---the part of $z$ the classifier discards---is \emph{structured} for ID data and \emph{disrupted} for OOD data (Figure~\ref{fig:method-hist}).

\paragraph{Observation.}
Within each class $c$, ID samples' residuals $z_\perp$ tend to cluster around a consistent direction.
Intuitively, samples of the same class share not only a similar projection onto the class weight (which determines their classification) but also similar ``leftover'' structure in the orthogonal complement---a class-specific signature that the classifier does not need for classification but that is characteristic of genuine class members.
OOD samples misclassified as class $c$ may match the projection (producing confident logits) but have arbitrary residual directions, since they were not drawn from the same data-generating process.

\paragraph{Remark: connection to Neural Collapse.}
The Neural Collapse phenomenon~\cite{papyan2020prevalence} predicts that well-trained classifiers drive within-class features toward their class mean along the classifier weight direction.
Convergence is imperfect: a structured residual persists in $z_\perp$, reflecting within-class variation consistent across genuine class members.
OOD inputs that mimic the class prototype along $w_c$ may exhibit comparably concentrated residuals, but oriented in a different direction that does not match the class-specific signature $\mu_\perp^{(c)}$.
We confirm this statistically across all five benchmark settings ($p < 0.01$, both CNNs and transformers; see supplementary).
This explains why the directional signature is class-specific: each class occupies a distinct region of the orthogonal complement, shaped by the class-specific data distribution rather than the classifier weights.

\paragraph{Offline estimation.}
Given a small set of ID training samples for each class $c$, we compute the mean residual direction:
\begin{equation}\label{eq:mu_perp}
    \mu_\perp^{(c)} = \text{normalize}\!\left(\frac{1}{|\mathcal{D}_c|} \sum_{x_i \in \mathcal{D}_c} z_{\perp,i}\right),
\end{equation}
where $\mathcal{D}_c$ is the set of training samples with label $c$, and $z_{\perp,i}$ is the orthogonal residual of $z_i = \phi(x_i)$ computed via Eq.~\eqref{eq:decomp}.

\paragraph{Online scoring.}
For a test sample with predicted class $\hat{y}$, the membership score is:
\begin{equation}\label{eq:membership}
    S_\text{mem}(x) = \cos(z_\perp,\, \mu_\perp^{(\hat{y})}) = \frac{z_\perp^\top \mu_\perp^{(\hat{y})}}{\|z_\perp\|}.
\end{equation}
ID samples yield high cosine similarity (their residuals align with the class-specific mean direction); OOD samples yield low similarity (their residuals point in arbitrary directions).

\subsection{Score Combination}\label{sec:combination}

The confidence and membership scores operate on different scales: Energy typically ranges over $[5, 30]$ while cosine similarity lies in $[-1, 1]$.
We normalize each using $z$-score statistics computed from the ID training set:
\begin{equation}\label{eq:normalize}
    S_\text{conf}' = \frac{S_\text{conf} - \mu_E}{\sigma_E}, \qquad
    S_\text{mem}' = \frac{S_\text{mem} - \mu_R}{\sigma_R},
\end{equation}
where $(\mu_E, \sigma_E)$ and $(\mu_R, \sigma_R)$ are the mean and standard deviation of each score on the ID training data.

The final \mine score is their sum:
\begin{equation}\label{eq:core}
    S_\text{\mine}(x) = S_\text{conf}' + S_\text{mem}'.
\end{equation}
\paragraph{Why summation.}
The sum provides OR semantics: a sample is flagged as OOD if \emph{either} its confidence is anomalously low \emph{or} its residual direction is anomalous.
This is the appropriate inductive bias because the two failure modes are complementary---confidence fails on OOD samples that produce high logits, while membership fails on OOD samples whose $z_\perp$ accidentally aligns with the ID direction.

\paragraph{Why principled.}
Because $z_\parallel \perp z_\perp$ by construction, the two scores are computed from orthogonal subspaces.
This orthogonality provides a stronger guarantee than ad hoc score fusion: the failure modes of the two components are approximately independent---not merely empirically uncorrelated but grounded in the geometric separation of the subspaces ($r{=}0.30$--$0.40$ on held-out data; \S\ref{sec:analysis}); under Neural Collapse geometry, where class weights converge to an equiangular tight frame, the cross-talk between $z_\perp$ and non-predicted class weights is further bounded.
We verify this empirically in \S\ref{sec:analysis}: the component scatter plots confirm that \mine's signals occupy distinct regions of the score space, unlike other score combination methods whose components cluster along a shared diagonal (Figure~\ref{fig:fusion-scatter}).

Like MDS~\cite{lee2018mahalanobis}, KNN~\cite{sun2022out}, and ViM~\cite{wang2022vim}, \mine requires a small set of ID training samples for offline estimation.
Our main experiments use only ${\sim}13$ samples per class for ImageNet calibration (${\sim}1\%$ of training data); \S\ref{sec:sample-efficiency} shows performance is stable from 1\% to 100\%.
Computational cost is analyzed in \S\ref{sec:cost}.

\section{Experiments}\label{sec:experiment}

\subsection{Setup}\label{sec:setup}

\paragraph{Benchmarks.}
We evaluate \mine across five benchmark settings spanning three CNN and two transformer architectures on two ID datasets:
CIFAR-100~\cite{krizhevsky2009learning} with ResNet-18~\cite{he2016deep} and WideResNet-40-2~\cite{zagoruyko2016wide}; and ImageNet~\cite{deng2009imagenet} with ResNet-50~\cite{he2016deep}, ViT-B/16~\cite{dosovitskiy2021image}, and Swin-B~\cite{liu2021swin}.
Each CIFAR-100 setting is evaluated against 8 OOD datasets and each ImageNet setting against 8, yielding 40 ID/OOD evaluation pairs in total.

\paragraph{Baselines.}
We compare against 16 post-hoc OOD scorers from five categories:
\emph{logit-based}---MSP~\cite{hendrycks2016baseline}, Energy~\cite{liu2020energy}, Max Logit~\cite{basart2022scaling};
\emph{feature-based}---KNN~\cite{sun2022out}, Mahalanobis~\cite{lee2018mahalanobis}, MDS++~\cite{ren2021simple}, CoP~\cite{fang2024cop}, CoRP~\cite{fang2024cop};
\emph{activation shaping}---ReAct~\cite{sun2021react}, ASH~\cite{djurisic2023ash}, SCALE~\cite{xu2024scale};
\emph{hybrid}---ViM~\cite{wang2022vim}, SHE~\cite{zhang2023she}, GradNorm~\cite{huang2021importance};
\emph{score combination}---NNGuide~\cite{park2023nnguide}, ComboOOD~\cite{rajasekaran2024combood}.

\paragraph{Metrics.}
We report AUROC~($\uparrow$) and FPR@95\%TPR~($\downarrow$), averaged across OOD datasets per setting.
All scorers are post-hoc (no retraining); features are cached once per model and shared across all methods.

\paragraph{Calibration data.}
Like KNN~\cite{sun2022out} and NNGuide~\cite{park2023nnguide}, several
methods require ID training samples for offline calibration (\eg fitting
class prototypes, $k$-NN indices, or normalization statistics).
For ImageNet, we provide a calibration set of ${\sim}1\%$ of training data; for CIFAR-100, the full training set. This lean budget tests robustness to limited calibration;
\S\ref{sec:sample-efficiency} confirms that \mine is stable across
budgets from 1\% to 100\%.
All methods that require calibration share the same calibration set per setting, ensuring a fair comparison.

\subsection{Main Results}\label{sec:main-results}

\begin{table}[!t]
\centering
\caption{Average AUROC~(\%)$\uparrow$\,/\,FPR@95~(\%)$\downarrow$ per setting. \textbf{Bold}: best. \underline{Underline}: second-best. L=Logit, F=Feature, S=Shaping, H=Hybrid, C=Combination.}
\label{tab:auroc}\label{tab:fpr}
{\footnotesize\setlength{\tabcolsep}{1.5pt}
\begin{tabular}{@{}l c cc ccc c@{}}
\toprule
& & \multicolumn{2}{c}{CIFAR-100} & \multicolumn{3}{c}{ImageNet} & \\
\cmidrule(lr){3-4}\cmidrule(lr){5-7}
Method & Cat. & R18 & WRN & R50 & ViT & Swin & Avg. \\
\midrule
MSP & L & 78.2/79.2 & 74.4/83.7 & 78.8/72.5 & 79.9/68.2 & 85.0/51.6 & 79.2/71.0 \\
Energy & L & 80.3/76.5 & 77.2/81.9 & 82.0/67.7 & 79.3/\underline{60.7} & 84.0/44.8 & 80.6/66.3 \\
Max Logit & L & 80.2/77.6 & 77.2/82.3 & 82.2/67.1 & 80.4/61.0 & 85.2/44.8 & 81.0/66.6 \\
\midrule
KNN & F & 81.5/71.9 & 80.8/\underline{70.6} & 73.0/69.5 & 44.8/97.7 & 83.6/68.6 & 72.7/75.6 \\
Mahalanobis & F & 66.3/89.7 & 77.5/76.3 & 63.6/83.6 & 77.2/84.1 & 87.9/53.8 & 74.5/77.5 \\
MDS++ & F & 81.1/\underline{71.0} & \underline{80.9}/73.2 & 82.5/58.2 & \underline{84.1}/63.6 & 89.3/45.0 & \underline{83.6}/62.2 \\
CoP & F & 71.4/79.7 & 57.4/90.5 & 76.0/66.0 & 68.1/85.8 & 88.2/45.8 & 72.2/73.5 \\
CoRP & F & 81.5/71.0 & 74.4/77.0 & 80.0/60.1 & 77.2/80.1 & 86.7/59.2 & 79.9/69.5 \\
\midrule
ReAct & S & 80.4/74.5 & 75.9/80.0 & 86.2/\textbf{49.9} & 81.8/\textbf{57.7} & 86.7/\underline{41.4} & 82.2/60.7 \\
ASH & S & 80.8/75.8 & 77.4/81.9 & 83.7/62.6 & 36.8/98.5 & 53.4/77.5 & 66.4/79.3 \\
SCALE & S & 81.4/74.1 & 78.1/78.7 & \underline{86.7}/52.0 & 71.0/72.8 & 85.0/44.8 & 80.4/64.5 \\
\midrule
ViM & H & 80.3/76.3 & 78.5/76.6 & 82.4/67.8 & 77.4/79.1 & \underline{89.6}/47.2 & 81.6/69.4 \\
SHE & H & 78.4/77.5 & 60.9/87.0 & 82.5/56.2 & 78.6/66.0 & 79.9/51.7 & 76.1/67.7 \\
GradNorm & H & 70.3/77.3 & 50.2/93.6 & 81.7/54.9 & 51.8/90.3 & 39.3/87.1 & 58.7/80.6 \\
\midrule
NNGuide & C & 81.5/74.6 & 79.0/77.7 & 85.9/\underline{50.7} & 80.0/74.8 & 88.8/45.1 & 83.0/64.6 \\
ComboOOD & C & \textbf{87.7}/\textbf{48.6} & \textbf{85.1}/\textbf{50.7} & 76.7/66.4 & 71.2/71.3 & 81.7/56.7 & 80.5/\textbf{58.7} \\
\midrule
\textbf{\mine} & C & \underline{82.1}/73.0 & 80.8/72.9 & \textbf{86.7}/51.2 & \textbf{84.3}/61.8 & \textbf{90.7}/\textbf{37.2} & \textbf{84.9}/\underline{59.2} \\
\bottomrule
\end{tabular}}
\end{table}

Table~\ref{tab:auroc} reports the average AUROC and FPR@95\%TPR across OOD datasets for each of the five settings.
Table~\ref{tab:per-dataset} breaks down two representative settings by individual OOD dataset, grouped into near-OOD (semantically close to the ID data) and far-OOD.

\begin{table}[!t]
\centering
\caption{Per-OOD-dataset AUROC~(\%) for two representative settings. \textbf{Bold}: best. \underline{Underline}: second-best. Near/far grouping follows OpenOOD~\cite{yang2022openood}.}
\label{tab:per-dataset}
{\footnotesize\setlength{\tabcolsep}{2.5pt}
\textit{(a) ResNet-18 / CIFAR-100}\\[2pt]
\begin{tabular}{@{}l cc cccccc c@{}}
\toprule
& \multicolumn{2}{c}{Near-OOD} & \multicolumn{6}{c}{Far-OOD} & \\
\cmidrule(lr){2-3}\cmidrule(lr){4-9}
Method & C-10 & TIN & SVHN & P365 & LSUN & iSUN & Text & MNIST & Avg. \\
\midrule
Energy & \underline{79.2} & 80.3 & 83.5 & 77.3 & 85.5 & 84.4 & 79.4 & 73.1 & 80.3 \\
KNN & 77.1 & \textbf{81.2} & 85.8 & 77.4 & 87.1 & 86.0 & 84.7 & 73.0 & 81.5 \\
ViM & 72.0 & 76.2 & 84.8 & 74.1 & 86.8 & 86.0 & 86.3 & 75.9 & 80.3 \\
MDS++ & 71.6 & 74.8 & \underline{89.7} & 75.1 & 86.2 & 86.2 & \underline{87.1} & \underline{78.4} & 81.1 \\
NNGuide & \textbf{79.5} & \underline{80.9} & 84.8 & \underline{77.8} & \underline{87.2} & \underline{86.4} & 82.2 & 73.2 & 81.5 \\
ComboOOD & 76.3 & 80.2 & \textbf{97.9} & 76.2 & \textbf{99.0} & \textbf{98.5} & \textbf{92.3} & \textbf{81.5} & \textbf{87.7} \\
\midrule
\textbf{\mine} & 77.4 & 79.3 & 89.5 & \textbf{78.1} & 85.6 & 85.2 & 85.1 & 76.2 & \underline{82.1} \\
\bottomrule
\end{tabular}\\[6pt]
\textit{(b) Swin-B / ImageNet}\\[2pt]
\begin{tabular}{@{}l ccc ccccc c@{}}
\toprule
& \multicolumn{3}{c}{Near-OOD} & \multicolumn{5}{c}{Far-OOD} & \\
\cmidrule(lr){2-4}\cmidrule(lr){5-9}
Method & SSB & NINCO & ImgO & Places & SUN & iNat & Text & OImgO & Avg. \\
\midrule
Energy & 79.1 & 84.4 & 82.8 & 77.0 & 81.3 & 95.2 & 81.8 & 90.7 & 84.0 \\
KNN & 72.1 & 76.8 & 84.9 & 81.2 & 84.5 & 91.5 & 90.0 & 88.1 & 83.6 \\
ViM & 79.3 & 88.1 & \underline{93.4} & \underline{84.3} & \underline{86.6} & \textbf{99.1} & 90.2 & 95.7 & \underline{89.6} \\
MDS++ & \underline{79.9} & \underline{90.9} & 93.3 & 80.8 & 82.9 & \underline{98.8} & 91.2 & \underline{96.3} & 89.3 \\
NNGuide & 77.3 & 88.1 & 93.0 & 82.4 & 84.7 & 98.1 & 91.0 & 96.0 & 88.8 \\
ComboOOD & 70.7 & 74.6 & \textbf{96.5} & 67.5 & 75.3 & 90.4 & \textbf{92.4} & 86.4 & 81.7 \\
\midrule
\textbf{\mine} & \textbf{83.1} & \textbf{90.9} & 92.6 & \textbf{85.0} & \textbf{87.9} & 98.6 & \underline{91.4} & \textbf{96.4} & \textbf{90.7} \\
\bottomrule
\end{tabular}}
\end{table}

\paragraph{Overall.}
\mine achieves the highest grand average AUROC (84.9\%), ahead of MDS++ (83.6\%) and NNGuide (83.0\%), and ranks \emph{first on all three ImageNet benchmarks}: R50 (86.7\%), ViT (84.3\%), Swin (90.7\%)---no other scorer achieves this.
The gain over Energy alone---which captures only the confidence signal that \mine also uses---is +4.7 (R50), +5.0 (ViT), and +6.7 (Swin)\,pp, demonstrating the large marginal value of the residual membership component on large-scale benchmarks.
On CIFAR-100, ComboOOD leads (87.7\% on R18, 85.1\% on WRN) but catastrophically drops on ImageNet transformers: $-$16.5\,pp on ViT.
Similarly, KNN drops $-$36.8\,pp and ASH drops $-$44.0\,pp from R18 to ViT.
\textbf{\mine never catastrophically fails} (minimum: 80.8\%), ranking first in three of five settings and remaining competitive in the other two.

\paragraph{Per-dataset breakdown.}
On Swin/ImageNet (Table~\ref{tab:per-dataset}b), \mine achieves the best AUROC on five of eight OOD datasets, with a notable near-OOD gain on SSB (+3.2\,pp vs.\ runner-up MDS++) and a tie on NINCO (90.9\%).
ComboOOD leads on ImageNet-O (96.5\%) and Textures (92.4\%), and ViM on iNaturalist (99.1\%); for \mine the gaps are small on these near-saturated benchmarks.
On R18/CIFAR-100 (Table~\ref{tab:per-dataset}a), NNGuide leads on the hardest near-OOD split (C-10: 79.5\%) while ComboOOD dominates far-OOD by a wide margin (e.g., LSUN: 99.0\%, SVHN: 97.9\%).
\mine is consistently competitive across both groups, ranking first on Places365 (78.1\%).
The 5.6\,pp gap to ComboOOD on R18 is driven entirely by far-OOD datasets (SVHN, LSUN, iSUN) where ComboOOD's Mahalanobis+$k$-NN combination excels on low-dimensional CNN features---the same components that collapse on ImageNet transformers.

\subsection{Ablation Study}\label{sec:ablation}

\begin{table}[!t]
\centering
\caption{Ablation studies (AUROC, \%). (a)~Component ablation: Confidence = Energy, Membership = $\cos(z_\perp, \mu_\perp^{(\hat{y})})$. (b)~Combination function: all variants use $z$-score normalization.}
\label{tab:ablation}
{\small\setlength{\tabcolsep}{4pt}
\begin{tabular}{@{}l cc ccc c@{}}
\toprule
& \multicolumn{2}{c}{CIFAR-100} & \multicolumn{3}{c}{ImageNet} & \\
\cmidrule(lr){2-3}\cmidrule(lr){4-6}
& R18 & WRN & R50 & ViT & Swin & Avg. \\
\midrule
\multicolumn{7}{@{}l}{\textit{(a) Component ablation}} \\
Confidence only & 80.3 & 77.2 & 82.0 & 79.3 & 84.0 & 80.6 \\
Membership only & 75.3 & 76.3 & 84.3 & 82.3 & 88.2 & 81.3 \\
\textbf{\mine} ($S_\text{conf}' {+} S_\text{mem}'$) & \textbf{82.1} & \textbf{80.8} & \textbf{86.7} & \textbf{84.3} & \textbf{90.7} & \textbf{84.9} \\
\midrule
\multicolumn{7}{@{}l}{\textit{(b) Combination function}} \\
\textbf{Sum} (OR) & \textbf{82.1} & \textbf{80.8} & \textbf{86.7} & \textbf{84.3} & \textbf{90.7} & \textbf{84.9} \\
Softmin ($\tau{=}5$) & 80.9 & 79.8 & 85.9 & 82.8 & 89.4 & 83.8 \\
Max & 79.9 & 79.0 & 84.2 & 83.3 & 87.3 & 82.7 \\
\bottomrule
\end{tabular}}
\end{table}

Table~\ref{tab:ablation}(a) shows that the two components are complementary: membership alone surpasses Energy on all three ImageNet settings (+2.3 to +4.2\,pp), confirming that $z_\perp$ carries substantial discriminative signal in high-dimensional feature spaces.
Conversely, confidence dominates on CIFAR-100 (+0.9 to +5.0\,pp) where shallower features (512--640\,dim) yield noisier residual estimates.
Their combination exceeds the best single component on all 5 settings, with the largest gain on WRN (+3.6\,pp) where both components are individually weak---precisely the setting where complementary fusion matters most.
Table~\ref{tab:ablation}(b) compares combination functions: summation (OR semantics) consistently outperforms softmin and max (84.9\% vs.\ 83.8\% and 82.7\%).

\subsection{Sample Efficiency}\label{sec:sample-efficiency}

\begin{table}[!t]
\centering
\caption{AUROC~(\%) vs.\ calibration budget for methods requiring initialization, averaged over 8 ImageNet OOD datasets.}
\label{tab:sample-eff}
{\small\setlength{\tabcolsep}{4pt}
\begin{tabular}{@{}l ccc c ccc c ccc c@{}}
\toprule
& \multicolumn{3}{c}{ResNet-50} && \multicolumn{3}{c}{ViT-B} && \multicolumn{3}{c}{Swin-B} \\
\cmidrule(lr){2-4}\cmidrule(lr){6-8}\cmidrule(lr){10-12}
Method & 1\% & 10\% & 100\% && 1\% & 10\% & 100\% && 1\% & 10\% & 100\% \\
\midrule
\textbf{\mine} & \textbf{86.7} & \textbf{87.4} & \textbf{87.5} && \textbf{84.3} & \textbf{85.1} & \textbf{85.1} && \textbf{90.7} & \textbf{90.9} & \textbf{91.0} \\
KNN & 73.0 & 73.3 & 79.5 && 44.8 & 37.0 & 77.8 && 83.6 & 71.4 & 85.9 \\
ViM & 82.4 & 83.0 & 83.0 && 77.4 & 78.0 & 78.1 && 89.6 & 90.4 & 90.4 \\
MDS++ & 82.5 & 83.9 & 83.9 && 84.1 & 84.8 & 84.8 && 89.3 & 90.1 & 90.2 \\
NNGuide & 85.9 & 85.3 & 84.6 && 80.0 & 80.6 & 79.5 && 88.8 & 89.7 & 89.5 \\
ComboOOD & 76.7 & 82.3 & 82.4 && 71.2 & 83.2 & 83.7 && 81.7 & 89.7 & 89.9 \\
\bottomrule
\end{tabular}}
\end{table}

Table~\ref{tab:sample-eff} varies the calibration budget from ${\sim}1\%$ (${\sim}13$ samples/class) to 100\% for all methods requiring initialization.
\mine's AUROC changes by at most 0.8\,pp between 1\% and 100\% (Swin: 0.3\,pp), because estimating a single mean direction $\mu_\perp^{(c)}$ converges far faster than fitting $d{\times}d$ covariance matrices (MDS++) or populating $k$-NN indices.
KNN is catastrophically data-sensitive: on ViT it plummets from 77.8\% (100\%) to 44.8\% (1\%), and on Swin it exhibits a non-monotonic U-shape (83.6\% at 1\%, 71.4\% at 10\%, 85.9\% at 100\%), suggesting that intermediate budgets can actually \emph{hurt} distance-based methods.
ComboOOD is similarly unstable, dropping from 83.7\% to 71.2\% on ViT at 1\%.
Even at 1\%, \mine's grand average across the three ImageNet settings (87.2\%) exceeds the \emph{100\%-data} grand average of every competitor (MDS++: 86.3\%, ComboOOD: 85.3\%, NNGuide: 84.5\%, ViM: 83.8\%, KNN: 81.1\%).
This confirms that the residual direction is an inherently low-dimensional statistic: a single unit vector per class suffices.

\subsection{Analysis: Why CORE Is Robust}\label{sec:analysis}

\begin{figure}[!t]
    \centering
    \includegraphics[width=\linewidth]{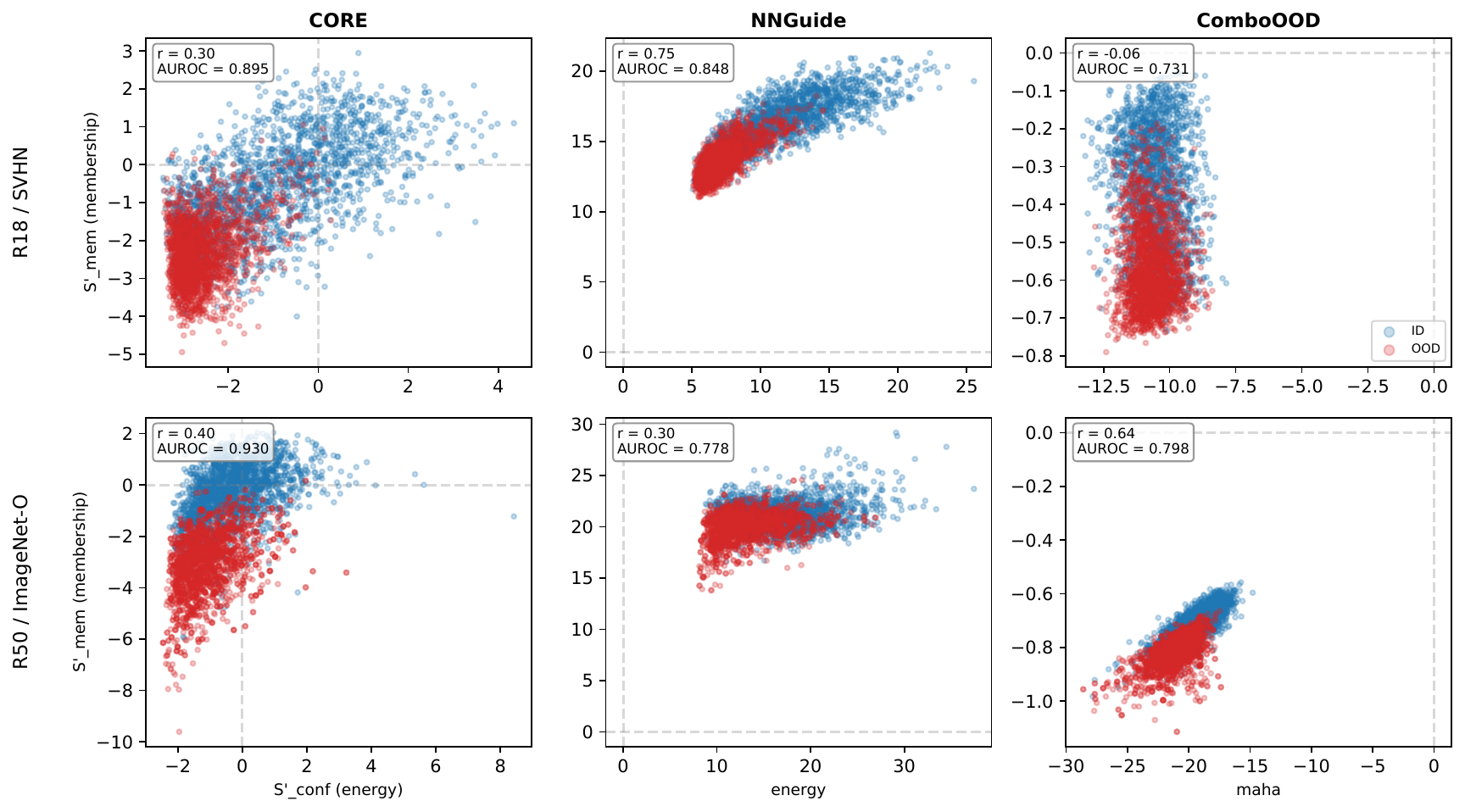}
    \caption{Component scatter plots for three score combination methods on two representative settings (rows: R18/SVHN, R50/ImageNet-O). \textbf{Left}: \mine's confidence ($S_\text{conf}'$) and membership ($S_\text{mem}'$) are weakly correlated ($r{=}0.30$, $0.40$); ID samples (blue) cluster at high values on both axes while OOD samples (red) scatter across complementary failure regions. \textbf{Center}: NNGuide's energy and $k$-NN signals are entangled ($r{=}0.75$ on R18) and overlap substantially. \textbf{Right}: ComboOOD's Mahalanobis component collapses on CNN features (ID and OOD overlap entirely on the $x$-axis), and its $k$-NN component inherits hubness on ImageNet.}
    \label{fig:fusion-scatter}
\end{figure}

Figure~\ref{fig:fusion-scatter} decomposes three score combination methods into their individual components.
\mine's confidence and membership signals are weakly correlated ($r{=}0.30$ on R18/SVHN, $0.40$ on R50/ImageNet-O): OOD samples that achieve high confidence but low membership---``confident misclassifications''---are caught by $S_\text{mem}'$, while the reverse case is caught by $S_\text{conf}'$.
This low correlation is precisely what makes the combination effective: when one signal is fooled, the other remains discriminative.
In contrast, NNGuide's two signals are structurally entangled ($r{=}0.75$ on R18) because both derive from the full feature~$z$, so the $k$-NN verification largely echoes what Energy already captures.
ComboOOD's components appear uncorrelated ($r{=}{-}0.06$ on R18), but this is an artifact: the Mahalanobis component collapses to chance-level discrimination (AUROC~$0.50$) on CNN features, contributing noise rather than signal; on ViT, $k$-NN collapses instead (AUROC~$0.36$) due to hubness.
\mine's orthogonal decomposition avoids both failure modes: each component operates on a distinct subspace, yielding complementary signals whose combination is robust across all five settings.

\subsection{Computational Cost}\label{sec:cost}

\mine adds negligible overhead: given the penultimate feature~$z$ and predicted class~$\hat{y}$, it requires two dot products ($z^\top w_{\hat{y}}$ and $z_\perp^\top \mu_\perp$), one vector subtraction, and two scalar $z$-score operations---$\mathcal{O}(d)$ per sample.
Storage consists of $\mu_\perp \in \mathbb{R}^{C \times d}$ and four normalization scalars $(\mu_E, \sigma_E, \mu_R, \sigma_R)$---for ImageNet ($C{=}1000$, $d{=}2048$), this is ${\sim}8$\,MB.
This is dramatically lighter than methods requiring $\mathcal{O}(d^2)$ covariance matrices (Mahalanobis, MDS++) or $\mathcal{O}(Nd)$ nearest-neighbor search (KNN, NNGuide).
In contrast, the nearest competing score combination method requires forward hooks on \emph{all} activation layers, a FAISS $k$-NN index over the full training set (${\sim}$10\,GB for ImageNet), and an inverse covariance matrix---orders of magnitude more compute and memory.

\section{Conclusion}\label{sec:conclusion}

We showed that the inconsistency of post-hoc OOD scorers stems from entangling confidence and membership in the full feature space.
\mine decomposes penultimate features into $z_\parallel$ (confidence) and $z_\perp$ (membership), combining two approximately independent signals via $z$-score summation.
Across five model$\times$dataset settings and 16 baselines, \mine achieves the highest grand average AUROC (84.9\%) at $\mathcal{O}(d)$ per-sample cost, with as few as ${\sim}1\%$ of training samples (\S\ref{sec:sample-efficiency}).
For practitioners, \mine offers a drop-in post-hoc scorer that requires no architecture-specific tuning, no covariance matrices, and no nearest-neighbor indices---only two dot products and one normalization per sample beyond a standard forward pass.

\paragraph{Limitations.}
\mine's membership signal degrades on semantically extremely adjacent OOD data (e.g., CIFAR-10 vs.\ CIFAR-100, $-$2\,pp vs.\ NNGuide) where $z_\perp$ shares the same directional structure as ID; this is inherent to any class-conditional membership method.

\bibliographystyle{splncs04}
\bibliography{main}

\clearpage
\appendix
\section{Supplementary Material}\label{sec:supplementary}

\subsection{Implementation Details}\label{sec:impl-details}

\paragraph{Architectures and layers.}
Table~\ref{tab:arch-details} lists the penultimate and logit layer names, feature dimensionality, and input resolution for each architecture.
All features are extracted as 1-D vectors after global average pooling (CNNs) or CLS token extraction (ViT) or spatial mean pooling (Swin).

\begin{table}[ht]
\centering
\caption{Architecture details.}
\label{tab:arch-details}
\small
\begin{tabular}{lllcc}
\toprule
Architecture & Penultimate & Logit & Dim & Size \\
\midrule
ResNet-18 & \texttt{avgpool} & \texttt{fc} & 512 & 32 \\
WRN-40-2 & \texttt{relu} & \texttt{fc} & 640 & 32 \\
ResNet-50 & \texttt{avgpool} & \texttt{fc} & 2048 & 224 \\
ViT-B/16 & \texttt{norm} / \texttt{vit.layernorm} & \texttt{head} / \texttt{classifier} & 768 & 224 \\
Swin-B & \texttt{swin.layernorm} & \texttt{classifier} & 1024 & 224 \\
\bottomrule
\end{tabular}
\end{table}

\paragraph{Model sources.}
\begin{itemize}
    \item \textbf{ResNet-18} (CIFAR-100): OpenOOD~\cite{zhang2023openood} pretrained checkpoint.
    \item \textbf{WRN-40-2} (CIFAR-100): RobustBench~\cite{croce2021robustbench} \texttt{Hendrycks2020AugMix\_WRN} with built-in normalization.
    \item \textbf{ResNet-50} (ImageNet): torchvision pretrained (ImageNet-1K).
    \item \textbf{ViT-B/16} (ImageNet): timm \texttt{vit\_base\_patch16\_224}; (CIFAR): HuggingFace \texttt{edumunozsala/vit\_base-224-in21k-ft-cifar100}.
    \item \textbf{Swin-B} (ImageNet): HuggingFace \texttt{microsoft/swin-base-patch4-window7-224}.
\end{itemize}

\paragraph{OOD datasets.}
For CIFAR-100 experiments: CIFAR-10, Tiny ImageNet (resized to 32$\times$32), SVHN, Places365, LSUN (resized), iSUN, DTD Textures, MNIST (converted to RGB).
For ImageNet experiments: SSB-Hard~\cite{vaze2022openset}, NINCO~\cite{bitterwolf2023ninco}, ImageNet-O~\cite{hendrycks2021nae}, Places~\cite{zhou2017places}, SUN~\cite{xiao2010sun}, iNaturalist~\cite{vanhorn2018inaturalist}, DTD Textures~\cite{cimpoi2014describing}, OpenImage-O~\cite{wang2022vim}.
Near/far-OOD groupings follow OpenOOD~\cite{yang2022openood}: for ImageNet, SSB-Hard, NINCO, and ImageNet-O are near-OOD; the remaining datasets are far-OOD.

\paragraph{Calibration data.}
For ImageNet settings, we use ${\sim}1\%$ of the training data (${\sim}$12{,}812 samples) as calibration data for all methods requiring initialization.
For CIFAR-100 settings, the full training set (50{,}000 samples) is used.
All methods that require calibration share the same calibration set per setting.

\subsection{Neural Collapse Verification}\label{sec:nc-verification}

We verify that the residual direction $z_\perp$ carries a genuine membership signal by measuring the alignment gap $\Delta = \mathbb{E}_\text{ID}[\cos(z_\perp, \mu_\perp^{(\hat{y})})] - \mathbb{E}_\text{OOD}[\cdot]$.
For each architecture, we compute class-specific mean residual directions $\mu_\perp^{(c)}$ from correctly-classified training samples, then score both ID test and OOD samples by $\cos(z_\perp, \mu_\perp^{(\hat{y})})$.
If Neural Collapse structures the orthogonal complement, ID residuals should align with $\mu_\perp$ while OOD residuals should not, yielding $\Delta > 0$.

Table~\ref{tab:nc-verification} confirms this across all five settings.
The 95\% confidence intervals are obtained via nonparametric bootstrap (10{,}000 resamples of the mean difference); $p$-values are from Welch's $t$-test (unequal-variance, one-sided $H_1{:}\,\mu_\text{ID} > \mu_\text{OOD}$).
All CIs exclude zero and all $p$-values are astronomically small, confirming that the gap is not a statistical artifact.
The gap is notably larger for transformers ($\Delta{=}0.267$ for ViT-B, $0.325$ for Swin-B) than for CNNs ($0.048$--$0.086$), suggesting that transformer features exhibit stronger class-specific structure in the orthogonal complement.

\begin{table}[t]
\centering
\caption{Neural Collapse verification: ID--OOD residual alignment gap $\Delta$ across architectures. All gaps are positive and statistically significant.}
\label{tab:nc-verification}
\small
\begin{tabular}{llccc}
\toprule
Setting & Architecture & $\Delta$ & 95\% CI & $p$-value \\
\midrule
CIFAR-100  & ResNet-18  & 0.057 & [0.052, 0.062] & $< 10^{-39}$ \\
CIFAR-100  & WRN        & 0.048 & [0.044, 0.053] & $< 10^{-40}$ \\
ImageNet   & ResNet-50  & 0.086 & [0.082, 0.090] & $< 10^{-152}$ \\
ImageNet   & ViT-B      & 0.267 & [0.255, 0.279] & $< 10^{-157}$ \\
ImageNet   & Swin-B     & 0.325 & [0.310, 0.339] & $< 10^{-152}$ \\
\bottomrule
\end{tabular}
\end{table}

\subsection{Choice of Confidence Score}\label{sec:conf-ablation}

\mine uses Energy~\cite{liu2020energy} as the confidence component $S_\text{conf}$.
Table~\ref{tab:conf-choice} replaces it with alternative logit-based scores to assess modularity.
The membership component $S_\text{mem} = \cos(z_\perp, \mu_\perp^{(\hat{y})})$ is fixed in all rows.

\begin{table}[t]
\centering
\caption{Ablation: choice of confidence score (AUROC, \%). $S_\text{mem}$ is the residual membership component, shared across all combined variants.}
\label{tab:conf-choice}
{\small\setlength{\tabcolsep}{4pt}
\begin{tabular}{@{}l ccccc c@{}}
\toprule
& \multicolumn{2}{c}{CIFAR-100} & \multicolumn{3}{c}{ImageNet} & \\
\cmidrule(lr){2-3}\cmidrule(lr){4-6}
Score & R18 & WRN & R50 & ViT & Swin & Avg. \\
\midrule
\multicolumn{7}{@{}l}{\textit{Confidence only (no $S_\text{mem}$)}} \\
MSP & 78.2 & 74.4 & 78.8 & 79.9 & 85.0 & 79.2 \\
Energy & 80.3 & 77.2 & 82.0 & 79.3 & 84.0 & 80.6 \\
Max Logit & 80.2 & 77.2 & 82.2 & 80.4 & 85.2 & 81.0 \\
\midrule
\multicolumn{7}{@{}l}{\textit{$S_\text{mem}$ only (no confidence)}} \\
$\cos(z_\perp, \mu_\perp)$ & 75.3 & 76.3 & 84.3 & 82.3 & 88.2 & 81.3 \\
\midrule
\multicolumn{7}{@{}l}{\textit{Confidence + $S_\text{mem}$ ($z$-score sum)}} \\
MSP + $S_\text{mem}$ & 78.6 & 77.3 & 85.0 & 82.9 & 88.9 & 82.5 \\
Max Logit + $S_\text{mem}$ & 82.1 & 80.7 & 86.7 & 84.2 & 90.2 & 84.8 \\
\textbf{Energy + $S_\text{mem}$ (\mine)} & \textbf{82.1} & \textbf{80.8} & \textbf{86.7} & \textbf{84.3} & \textbf{90.7} & \textbf{84.9} \\
\bottomrule
\end{tabular}}
\end{table}

Energy is the best confidence component on all five settings, though Max Logit~+~$S_\text{mem}$ performs comparably (84.8\% vs.\ 84.9\%).
All three confidence scores improve substantially when combined with $S_\text{mem}$: MSP gains +0.4 to +6.2\,pp, Max Logit +1.9 to +5.0\,pp, confirming that the membership signal is complementary to \emph{any} logit-based score.
Energy is preferred as it marginally outperforms alternatives and is the most widely adopted confidence baseline~\cite{liu2020energy}.

\subsection{Component Correlation Analysis}\label{sec:correlation}

Table~\ref{tab:correlation} reports the Pearson correlation $r$ between the $z$-score-normalized confidence $S_\text{conf}'$ and membership $S_\text{mem}'$ components of \mine across all five settings.
On four of five settings, $|r| \leq 0.32$, confirming that the two signals are approximately independent.
The exception is WRN ($r_\text{ID}{=}0.49$), whose wider architecture produces features with higher confidence--membership redundancy.
This directly explains \mine's weakest relative ranking on WRN (3rd, behind ComboOOD and MDS++): when $S_\text{conf}$ and $S_\text{mem}$ are correlated, their combination yields less complementary information, reducing the benefit of the dual-score design.
Conversely, on transformers (ViT, Swin) where $r < 0.10$, \mine achieves its largest gains over single-signal baselines.

\begin{table}[t]
\centering
\caption{Pearson correlation between $S_\text{conf}'$ and $S_\text{mem}'$.}
\label{tab:correlation}
\small
\begin{tabular}{llcc}
\toprule
Setting & Architecture & $r$ (ID) & $r$ (OOD avg.) \\
\midrule
CIFAR-100 & ResNet-18 & 0.08 & 0.07 \\
CIFAR-100 & WRN-40-2 & 0.49 & 0.30 \\
ImageNet & ResNet-50 & 0.32 & 0.15 \\
ImageNet & ViT-B & 0.07 & 0.10 \\
ImageNet & Swin-B & 0.06 & 0.05 \\
\bottomrule
\end{tabular}
\end{table}

\subsection{Sensitivity to Hyperparameters}\label{sec:sensitivity}

\mine has two design choices beyond the method itself: normalization mode, and the combination weight~$\alpha$ in $S = \alpha \, S_\text{conf}' + (1{-}\alpha) \, S_\text{mem}'$.
We ablate each while fixing the others at their defaults ($z$-score, $\alpha{=}0.5$).

\paragraph{Normalization mode.}
Table~\ref{tab:norm-ablation} compares $z$-score, min-max, and no normalization.
$z$-score and min-max perform within 0.2\,pp of each other (84.9 vs.\ 84.7), while omitting normalization drops performance by 2.8\,pp.
Normalization matters---but the specific choice does not.

\begin{table}[t]
\centering
\caption{Normalization mode ablation (AUROC, \%).}
\label{tab:norm-ablation}
{\small\setlength{\tabcolsep}{4pt}
\begin{tabular}{@{}l ccccc c@{}}
\toprule
& \multicolumn{2}{c}{CIFAR-100} & \multicolumn{3}{c}{ImageNet} & \\
\cmidrule(lr){2-3}\cmidrule(lr){4-6}
Normalization & R18 & WRN & R50 & ViT & Swin & Avg. \\
\midrule
$z$-score & \textbf{82.1} & \textbf{80.8} & 86.7 & \textbf{84.3} & \textbf{90.7} & \textbf{84.9} \\
min-max & 81.5 & \textbf{80.8} & \textbf{86.8} & 83.9 & 90.6 & 84.7 \\
none & 80.5 & 77.4 & 82.3 & 82.3 & 88.1 & 82.1 \\
\bottomrule
\end{tabular}}
\end{table}

\begin{figure}[t]
\centering
\includegraphics[width=0.75\linewidth]{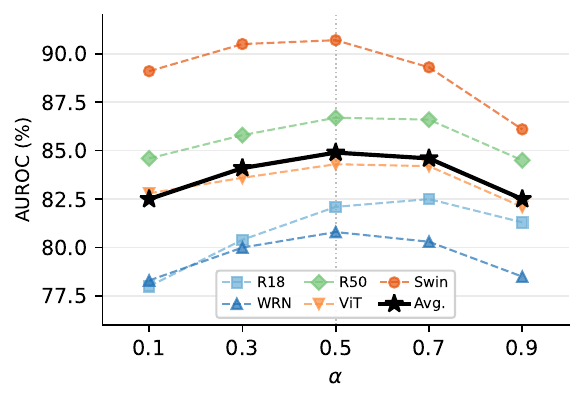}
\caption{Combination weight $\alpha$ ablation. $S = \alpha \, S_\text{conf}' + (1{-}\alpha) \, S_\text{mem}'$. All settings peak near $\alpha{=}0.5$ with a wide plateau.}
\label{fig:alpha-ablation}
\end{figure}

\paragraph{Combination weight $\alpha$.}
Fig.~\ref{fig:alpha-ablation} sweeps $\alpha$ from 0.1 (membership-dominated) to 0.9 (confidence-dominated).
Performance is remarkably flat across $\alpha \in [0.3, 0.7]$ (84.1--84.9\,\%), with $\alpha{=}0.5$ at the peak.
Only the extremes ($\alpha{=}0.1$ or $0.9$) degrade substantially ($-$2.4\,pp), confirming that equal weighting is a robust default requiring no tuning.

\subsection{Computational Cost}\label{sec:suppl-cost}

Table~\ref{tab:cost} compares the online scoring cost and offline storage of \mine against representative baselines on ResNet-50/ImageNet ($d{=}2048$, $C{=}1000$).
All timings are measured on a single NVIDIA A6000.

\begin{table}[t]
\centering
\caption{Computational cost comparison (ResNet-50 / ImageNet).}
\label{tab:cost}
{\small\setlength{\tabcolsep}{3pt}
\begin{tabular}{@{}llccl@{}}
\toprule
Method & Complexity & Latency & Storage & Init \\
       &            & (ms/samp) & (MB) & \\
\midrule
MSP / Energy      & $\mathcal{O}(C)$   & ${<}0.001$ & 0   & --- \\
ReAct / ASH / SCALE & $\mathcal{O}(Cd)$ & ${<}0.001$ & ${<}0.1$ & Thr. \\
SHE               & $\mathcal{O}(Cd)$  & ${<}0.01$  & 8   & $\mu_c$ \\
ViM               & $\mathcal{O}(d^2)$ & ${\sim}0.01$ & 8 & Null sp. \\
Maha.\ / MDS++    & $\mathcal{O}(d^2)$ & ${\sim}0.01$ & 32 & $\Sigma^{-1}$ \\
CoP               & $\mathcal{O}(dq)$  & ${<}0.01$  & 12  & PCA \\
CoRP              & $\mathcal{O}(dM)$  & ${\sim}0.05$ & 80 & RFF+PCA \\
KNN / NNGuide     & $\mathcal{O}(Nd)$  & 0.34       & 105 & Train ft. \\
GradNorm          & $\mathcal{O}(Cd){+}$bkwd & ${\sim}1$ & 0 & --- \\
\midrule
\textbf{\mine}    & $\mathcal{O}(d)$   & \textbf{${<}0.001$} & 8 & $\mu_\perp,\mu/\sigma$ \\
\bottomrule
\end{tabular}}
\end{table}

\mine's online scoring requires only two dot products (projection + cosine), one subtraction, and two scalar operations per sample---$\mathcal{O}(d)$, the lowest complexity among all methods that use features.
Its storage (8\,MB) consists of per-class residual directions $\mu_\perp^{(c)} \in \mathbb{R}^{C \times d}$ and four normalization scalars, comparable to methods like SHE while substantially cheaper than KNN/NNGuide (105\,MB train features) or Mahalanobis ($d \times d$ precision matrix).
Offline initialisation computes $\mu_\perp$ and $z$-score statistics from the calibration set in $< 0.02$\,s---orders of magnitude faster than KNN (index construction) or MDS (covariance inversion).

\subsection{Full Per-Dataset Results}\label{sec:full-results}

The main paper (Table~2) reports per-OOD-dataset breakdowns for two representative settings (ResNet-18/CIFAR-100 and Swin-B/ImageNet).
Here we provide the complete per-dataset AUROC~(\%) and FPR@95~(\%) for all five settings.
Near/far-OOD groupings follow OpenOOD~\cite{yang2022openood}: for ImageNet, SSB-Hard, NINCO, and ImageNet-O are near-OOD; the remaining datasets are far-OOD.
\textbf{Bold}: best; \underline{underline}: second-best.

\begin{table}[t]
\centering
\caption{Per-OOD-dataset AUROC~(\%)$\uparrow$ --- ResNet-18 / CIFAR-100.}
\label{tab:full-r18-auroc}
{\footnotesize\setlength{\tabcolsep}{2.5pt}
\begin{tabular}{@{}l cc cccccc c@{}}
\toprule
& \multicolumn{2}{c}{Near-OOD} & \multicolumn{6}{c}{Far-OOD} & \\
\cmidrule(lr){2-3}\cmidrule(lr){4-9}
Method & C-10 & TIN & SVHN & P365 & LSUN & iSUN & Text & MNIST & Avg. \\
\midrule
MSP & 78.7 & 79.9 & 80.3 & 77.4 & 82.1 & 81.3 & 78.2 & 68.0 & 78.2 \\
Energy & 79.2 & 80.3 & 83.5 & 77.3 & 85.5 & 84.4 & 79.4 & 73.1 & 80.3 \\
Max Logit & 79.3 & 80.5 & 83.2 & 77.5 & 85.0 & 84.0 & 79.4 & 72.5 & 80.2 \\
KNN & 77.1 & \textbf{81.2} & 85.8 & 77.4 & 87.1 & 86.0 & 84.7 & 73.0 & 81.5 \\
Mahalanobis & 55.8 & 61.2 & 72.8 & 62.7 & 69.8 & 70.1 & 76.3 & 61.7 & 66.3 \\
MDS++ & 71.6 & 74.8 & 89.7 & 75.1 & 86.2 & 86.2 & \underline{87.1} & 78.4 & 81.1 \\
CoP & 57.7 & 61.2 & 82.9 & 61.1 & 74.4 & 75.3 & 80.4 & 78.0 & 71.4 \\
CoRP & 70.6 & 74.3 & \underline{91.0} & 73.8 & 87.0 & 86.5 & 87.0 & \textbf{81.6} & 81.5 \\
ReAct & 75.4 & 79.2 & 82.6 & \textbf{78.6} & \underline{88.6} & \underline{88.0} & 85.7 & 65.0 & 80.4 \\
ASH & 79.3 & 80.3 & 84.4 & 77.5 & 85.8 & 84.8 & 80.1 & 73.8 & 80.8 \\
SCALE & \underline{79.4} & 80.2 & 86.6 & \underline{78.2} & 85.4 & 84.9 & 81.4 & 75.1 & 81.4 \\
ViM & 72.0 & 76.2 & 84.8 & 74.1 & 86.8 & 86.0 & 86.3 & 75.9 & 80.3 \\
SHE & 78.3 & 77.2 & 84.3 & 74.5 & 83.0 & 82.0 & 75.0 & 73.0 & 78.4 \\
GradNorm & 70.5 & 67.2 & 80.0 & 68.2 & 72.6 & 72.5 & 64.9 & 66.8 & 70.3 \\
NNGuide & \textbf{79.5} & \underline{80.9} & 84.8 & 77.8 & 87.2 & 86.4 & 82.2 & 73.2 & 81.5 \\
ComboOOD & 76.3 & 80.2 & \textbf{97.9} & 76.2 & \textbf{99.0} & \textbf{98.5} & \textbf{92.3} & \underline{81.5} & \textbf{87.7} \\
\midrule
\textbf{\mine} & 77.4 & 79.3 & 89.5 & 78.1 & 85.6 & 85.2 & 85.1 & 76.2 & \underline{82.1} \\
\bottomrule
\end{tabular}}
\end{table}

\begin{table}[t]
\centering
\caption{Per-OOD-dataset FPR@95\,(\%)$\downarrow$ --- ResNet-18 / CIFAR-100.}
\label{tab:full-r18-fpr}
{\footnotesize\setlength{\tabcolsep}{2.5pt}
\begin{tabular}{@{}l cc cccccc c@{}}
\toprule
& \multicolumn{2}{c}{Near-OOD} & \multicolumn{6}{c}{Far-OOD} & \\
\cmidrule(lr){2-3}\cmidrule(lr){4-9}
Method & C-10 & TIN & SVHN & P365 & LSUN & iSUN & Text & MNIST & Avg. \\
\midrule
MSP & 79.9 & 75.7 & 79.4 & 80.0 & 73.0 & 75.5 & 79.7 & 90.5 & 79.2 \\
Energy & 79.6 & \textbf{74.9} & 76.6 & 80.4 & 66.0 & 68.6 & 77.4 & 88.4 & 76.5 \\
Max Logit & \underline{79.3} & 75.4 & 77.6 & 80.0 & 68.9 & 71.5 & 78.5 & 89.4 & 77.6 \\
KNN & 80.9 & \underline{74.9} & 62.4 & \underline{78.2} & 61.6 & 66.0 & 63.6 & 88.1 & 71.9 \\
Mahalanobis & 95.8 & 94.4 & 91.5 & 93.2 & 87.4 & 84.6 & 73.4 & 97.6 & 89.7 \\
MDS++ & 86.8 & 85.4 & 51.6 & 81.3 & 62.8 & \underline{60.3} & \underline{53.8} & 85.9 & \underline{71.0} \\
CoP & 92.0 & 91.4 & 68.7 & 91.0 & 77.9 & 73.1 & 61.9 & 81.2 & 79.7 \\
CoRP & 88.4 & 86.8 & \underline{48.7} & 84.8 & 65.1 & 63.5 & 55.3 & \textbf{75.1} & 71.0 \\
ReAct & 83.7 & 76.8 & 77.1 & 80.1 & \underline{59.8} & 61.3 & 67.0 & 90.4 & 74.5 \\
ASH & 79.5 & 75.3 & 73.5 & 80.3 & 65.3 & 67.9 & 76.9 & 87.5 & 75.8 \\
SCALE & \textbf{78.9} & 75.4 & 65.7 & \textbf{78.1} & 66.8 & 68.8 & 73.8 & 85.5 & 74.1 \\
ViM & 87.8 & 85.1 & 74.2 & 84.0 & 64.8 & 64.5 & 59.5 & 90.5 & 76.3 \\
SHE & 80.0 & 80.2 & 68.0 & 82.5 & 72.2 & 72.9 & 80.4 & 84.1 & 77.5 \\
GradNorm & 80.6 & 79.7 & 70.2 & 81.8 & 71.5 & 71.9 & 79.0 & 84.0 & 77.3 \\
NNGuide & 80.8 & 75.3 & 72.8 & 80.1 & 61.3 & 64.4 & 73.1 & 88.8 & 74.6 \\
ComboOOD & 89.1 & 76.0 & \textbf{11.4} & 80.2 & \textbf{5.2} & \textbf{7.6} & \textbf{38.6} & \underline{80.5} & \textbf{48.6} \\
\midrule
\textbf{\mine} & 83.7 & 82.2 & 50.4 & 80.2 & 70.5 & 70.5 & 64.2 & 82.6 & 73.0 \\
\bottomrule
\end{tabular}}
\end{table}

\begin{table}[t]
\centering
\caption{Per-OOD-dataset AUROC~(\%)$\uparrow$ --- WRN / CIFAR-100.}
\label{tab:full-wrn-auroc}
{\footnotesize\setlength{\tabcolsep}{2.5pt}
\begin{tabular}{@{}l cc cccccc c@{}}
\toprule
& \multicolumn{2}{c}{Near-OOD} & \multicolumn{6}{c}{Far-OOD} & \\
\cmidrule(lr){2-3}\cmidrule(lr){4-9}
Method & C-10 & TIN & SVHN & P365 & LSUN & iSUN & Text & MNIST & Avg. \\
\midrule
MSP & 76.6 & 76.7 & 77.2 & 75.0 & 76.8 & 74.5 & 72.7 & 65.4 & 74.4 \\
Energy & \underline{78.7} & 77.8 & 78.7 & 75.7 & 78.6 & 74.6 & 76.3 & 76.8 & 77.2 \\
Max Logit & \textbf{78.9} & 78.1 & 79.1 & 76.0 & 78.9 & 75.0 & 76.3 & 75.4 & 77.2 \\
KNN & 76.1 & \textbf{79.7} & \underline{90.7} & 74.7 & 85.1 & 82.2 & 87.8 & 70.3 & 80.8 \\
Mahalanobis & 67.9 & 75.3 & 87.2 & 72.5 & \underline{86.1} & \underline{85.0} & 85.1 & 60.9 & 77.5 \\
MDS++ & 75.8 & 77.1 & 89.3 & 74.5 & 80.1 & 78.9 & \underline{90.6} & 80.7 & \underline{80.9} \\
CoP & 49.6 & 48.5 & 63.0 & 44.5 & 49.4 & 52.4 & 77.5 & 74.2 & 57.4 \\
CoRP & 64.4 & 66.9 & 89.1 & 61.8 & 70.4 & 71.5 & 90.0 & 80.8 & 74.4 \\
ReAct & 70.0 & 75.9 & 80.2 & \textbf{79.0} & 82.5 & 80.4 & 83.3 & 55.7 & 75.9 \\
ASH & 78.6 & 77.4 & 79.6 & 75.1 & 77.1 & 73.2 & 78.8 & 79.6 & 77.4 \\
SCALE & 77.8 & 76.4 & 81.3 & 75.6 & 75.1 & 72.7 & 83.0 & 82.6 & 78.1 \\
ViM & 72.3 & 75.4 & 84.9 & 71.0 & 81.4 & 80.4 & 88.6 & 73.9 & 78.5 \\
SHE & 66.4 & 56.8 & 65.6 & 56.3 & 38.9 & 38.7 & 74.3 & \textbf{89.8} & 60.9 \\
GradNorm & 55.1 & 46.2 & 46.1 & 48.7 & 28.1 & 29.0 & 64.8 & \underline{83.8} & 50.2 \\
NNGuide & 78.2 & \underline{78.2} & 83.7 & \underline{76.8} & 79.4 & 76.6 & 81.1 & 78.2 & 79.0 \\
ComboOOD & 70.2 & 74.8 & \textbf{95.7} & 72.1 & \textbf{99.7} & \textbf{99.1} & \textbf{94.1} & 75.2 & \textbf{85.1} \\
\midrule
\textbf{\mine} & 76.9 & 78.2 & 88.5 & 75.4 & 78.6 & 76.7 & 90.1 & 81.8 & 80.8 \\
\bottomrule
\end{tabular}}
\end{table}

\begin{table}[t]
\centering
\caption{Per-OOD-dataset FPR@95\,(\%)$\downarrow$ --- WRN / CIFAR-100.}
\label{tab:full-wrn-fpr}
{\footnotesize\setlength{\tabcolsep}{2.5pt}
\begin{tabular}{@{}l cc cccccc c@{}}
\toprule
& \multicolumn{2}{c}{Near-OOD} & \multicolumn{6}{c}{Far-OOD} & \\
\cmidrule(lr){2-3}\cmidrule(lr){4-9}
Method & C-10 & TIN & SVHN & P365 & LSUN & iSUN & Text & MNIST & Avg. \\
\midrule
MSP & 80.7 & 80.7 & 82.9 & 83.0 & 81.2 & 84.2 & 85.2 & 91.5 & 83.7 \\
Energy & \underline{79.1} & 78.4 & 84.8 & 81.2 & 82.2 & 86.6 & 82.8 & 79.7 & 81.9 \\
Max Logit & 79.5 & 78.7 & 83.9 & 81.7 & 81.4 & 86.1 & 83.4 & 83.5 & 82.3 \\
KNN & 82.7 & \textbf{75.7} & \underline{45.8} & 81.2 & 69.7 & 71.8 & 51.0 & 86.5 & \underline{70.6} \\
Mahalanobis & 91.3 & 84.2 & 61.1 & 85.9 & \underline{64.4} & \underline{62.6} & 61.6 & 99.0 & 76.3 \\
MDS++ & 84.0 & 81.6 & 56.9 & 82.8 & 83.3 & 82.3 & 43.4 & 71.7 & 73.2 \\
CoP & 96.5 & 96.7 & 92.9 & 96.9 & 97.5 & 95.6 & 61.4 & 86.5 & 90.5 \\
CoRP & 91.5 & 89.1 & 54.0 & 91.5 & 89.6 & 86.8 & \underline{37.0} & 76.8 & 77.0 \\
ReAct & 88.0 & 79.7 & 79.5 & \textbf{77.3} & 73.2 & 77.2 & 70.4 & 95.0 & 80.0 \\
ASH & 79.1 & 80.1 & 83.5 & 82.9 & 84.9 & 88.8 & 79.5 & 76.6 & 81.9 \\
SCALE & \textbf{78.6} & 79.1 & 77.8 & 81.4 & 84.0 & 87.0 & 68.7 & 73.2 & 78.7 \\
ViM & 86.9 & 82.6 & 67.2 & 86.0 & 75.0 & 72.9 & 50.6 & 91.3 & 76.6 \\
SHE & 88.6 & 93.8 & 90.8 & 95.1 & 98.9 & 98.6 & 72.5 & \textbf{58.0} & 87.0 \\
GradNorm & 94.0 & 97.0 & 98.4 & 97.0 & 99.4 & 99.3 & 84.9 & 78.8 & 93.6 \\
NNGuide & 79.2 & \underline{76.8} & 73.7 & \underline{79.7} & 79.2 & 83.6 & 72.6 & 76.8 & 77.7 \\
ComboOOD & 91.6 & 84.1 & \textbf{22.6} & 83.8 & \textbf{1.7} & \textbf{3.5} & \textbf{28.0} & 90.0 & \textbf{50.7} \\
\midrule
\textbf{\mine} & 83.0 & 79.6 & 57.8 & 82.6 & 84.8 & 84.6 & 44.5 & \underline{66.1} & 72.9 \\
\bottomrule
\end{tabular}}
\end{table}

\begin{table}[t]
\centering
\caption{Per-OOD-dataset AUROC~(\%)$\uparrow$ --- ResNet-50 / ImageNet.}
\label{tab:full-r50-auroc}
{\footnotesize\setlength{\tabcolsep}{2.5pt}
\begin{tabular}{@{}l ccc ccccc c@{}}
\toprule
& \multicolumn{3}{c}{Near-OOD} & \multicolumn{5}{c}{Far-OOD} & \\
\cmidrule(lr){2-4}\cmidrule(lr){5-9}
Method & SSB & NINCO & ImgO & Places & SUN & iNat & Text & OImgO & Avg. \\
\midrule
MSP & 70.7 & 79.9 & 69.8 & 79.0 & 80.1 & 88.2 & 78.9 & 83.8 & 78.8 \\
Energy & 70.3 & 79.5 & 75.0 & 82.1 & 84.9 & 91.3 & 84.9 & 88.3 & 82.0 \\
Max Logit & 70.7 & 80.3 & 74.9 & 82.2 & 84.8 & 91.6 & 84.6 & 88.3 & 82.2 \\
KNN & 49.3 & 60.2 & 89.1 & 64.7 & 72.2 & 84.1 & 92.8 & 71.2 & 73.0 \\
Mahalanobis & 46.2 & 60.2 & 86.0 & 51.1 & 49.8 & 57.6 & 91.1 & 66.5 & 63.6 \\
MDS++ & 59.0 & 77.0 & \underline{95.3} & 73.6 & 79.8 & 88.4 & \textbf{98.6} & 88.3 & 82.5 \\
CoP & 50.9 & 65.7 & 93.0 & 65.6 & 73.7 & 80.8 & \underline{97.8} & 80.9 & 76.0 \\
CoRP & 53.9 & 71.4 & \textbf{96.0} & 71.9 & 78.6 & 86.6 & 97.8 & 83.4 & 80.0 \\
ReAct & \underline{71.6} & 80.0 & 77.3 & \textbf{91.6} & \textbf{94.1} & \textbf{96.5} & 88.6 & 89.4 & 86.2 \\
ASH & 69.4 & 80.5 & 78.6 & 83.4 & 86.4 & 93.0 & 88.1 & 89.9 & 83.7 \\
SCALE & \textbf{72.0} & \textbf{83.6} & 82.4 & \underline{86.4} & \underline{89.4} & \underline{96.0} & 91.5 & \textbf{92.0} & \underline{86.7} \\
ViM & 63.1 & 78.5 & 91.4 & 76.2 & 78.4 & 86.5 & 96.9 & 88.3 & 82.4 \\
SHE & 68.3 & 75.5 & 79.5 & 81.9 & 86.3 & 92.8 & 90.3 & 85.1 & 82.5 \\
GradNorm & 67.2 & 73.2 & 78.6 & 82.2 & 87.5 & 94.1 & 87.9 & 83.1 & 81.7 \\
NNGuide & 70.9 & 82.2 & 81.0 & 85.9 & 88.9 & 95.5 & 91.9 & 91.0 & 85.9 \\
ComboOOD & 51.8 & 66.0 & 94.5 & 66.9 & 74.8 & 84.0 & 96.4 & 79.0 & 76.7 \\
\midrule
\textbf{\mine} & 67.6 & \underline{82.5} & 92.9 & 82.3 & 86.4 & 95.2 & 95.4 & \underline{91.1} & \textbf{86.7} \\
\bottomrule
\end{tabular}}
\end{table}

\begin{table}[t]
\centering
\caption{Per-OOD-dataset FPR@95\,(\%)$\downarrow$ --- ResNet-50 / ImageNet.}
\label{tab:full-r50-fpr}
{\footnotesize\setlength{\tabcolsep}{2.5pt}
\begin{tabular}{@{}l ccc ccccc c@{}}
\toprule
& \multicolumn{3}{c}{Near-OOD} & \multicolumn{5}{c}{Far-OOD} & \\
\cmidrule(lr){2-4}\cmidrule(lr){5-9}
Method & SSB & NINCO & ImgO & Places & SUN & iNat & Text & OImgO & Avg. \\
\midrule
MSP & 85.9 & 76.6 & 80.3 & 75.6 & 72.9 & 52.4 & 70.1 & 66.3 & 72.5 \\
Energy & 85.6 & 76.8 & 77.4 & 70.1 & 64.2 & 49.5 & 57.9 & 59.9 & 67.7 \\
Max Logit & 85.1 & 75.2 & 77.2 & 69.4 & 65.0 & 46.5 & 59.2 & 59.2 & 67.1 \\
KNN & 94.3 & 89.4 & 52.6 & 84.7 & 76.0 & 56.8 & 29.6 & 72.3 & 69.5 \\
Mahalanobis & 97.1 & 95.5 & 61.6 & 97.3 & 97.1 & 95.2 & 38.0 & 86.9 & 83.6 \\
MDS++ & 91.4 & 77.8 & \underline{27.1} & 80.8 & 72.5 & 57.1 & \textbf{6.9} & 52.0 & 58.2 \\
CoP & 93.5 & 84.8 & 40.3 & 87.0 & 79.8 & 69.5 & \underline{10.6} & 62.7 & 66.0 \\
CoRP & 91.9 & 82.3 & \textbf{23.8} & 80.8 & 71.6 & 59.4 & 10.9 & 60.3 & 60.1 \\
ReAct & \textbf{79.4} & 70.2 & 70.9 & \textbf{35.7} & \textbf{26.0} & \textbf{18.3} & 51.9 & 46.5 & \textbf{49.9} \\
ASH & 86.7 & 74.5 & 71.7 & 67.4 & 59.7 & 40.0 & 47.6 & 53.2 & 62.6 \\
SCALE & 83.2 & \underline{67.3} & 62.6 & 57.0 & 47.9 & \underline{22.1} & 34.1 & \textbf{41.7} & 52.0 \\
ViM & 92.4 & 81.9 & 44.5 & 85.0 & 85.3 & 76.4 & 14.7 & 62.4 & 67.8 \\
SHE & 85.2 & 72.6 & 64.5 & 60.6 & 49.1 & 31.1 & 34.3 & 52.0 & 56.2 \\
GradNorm & 84.5 & 73.8 & 66.5 & 56.6 & \underline{43.7} & 24.5 & 38.9 & 50.6 & 54.9 \\
NNGuide & \underline{82.4} & \textbf{67.2} & 61.6 & \underline{55.6} & 45.2 & 22.1 & 30.0 & \underline{41.9} & \underline{50.7} \\
ComboOOD & 93.5 & 87.1 & 34.4 & 85.2 & 76.8 & 64.2 & 18.2 & 71.7 & 66.4 \\
\midrule
\textbf{\mine} & 86.7 & 69.9 & 35.1 & 67.0 & 57.4 & 26.8 & 19.8 & 46.8 & 51.2 \\
\bottomrule
\end{tabular}}
\end{table}

\begin{table}[t]
\centering
\caption{Per-OOD-dataset AUROC~(\%)$\uparrow$ --- ViT-B / ImageNet.}
\label{tab:full-vit-auroc}
{\footnotesize\setlength{\tabcolsep}{2.5pt}
\begin{tabular}{@{}l ccc ccccc c@{}}
\toprule
& \multicolumn{3}{c}{Near-OOD} & \multicolumn{5}{c}{Far-OOD} & \\
\cmidrule(lr){2-4}\cmidrule(lr){5-9}
Method & SSB & NINCO & ImgO & Places & SUN & iNat & Text & OImgO & Avg. \\
\midrule
MSP & 69.1 & 79.4 & 78.8 & 79.7 & 81.2 & 88.7 & 77.7 & 84.4 & 79.9 \\
Energy & \underline{71.5} & 77.8 & 78.9 & 77.1 & 81.3 & 87.4 & 79.4 & 81.2 & 79.3 \\
Max Logit & 71.0 & 79.4 & 79.6 & 79.0 & 82.0 & 89.3 & 79.2 & 83.4 & 80.4 \\
KNN & 43.4 & 42.6 & 36.9 & 42.3 & 44.0 & 50.2 & 54.6 & 44.1 & 44.8 \\
Mahalanobis & 62.3 & 76.1 & 81.8 & 70.9 & 72.0 & 90.2 & 77.1 & 87.0 & 77.2 \\
MDS++ & 71.0 & \textbf{84.5} & 85.5 & 77.6 & 80.3 & \textbf{95.3} & \underline{86.7} & \textbf{91.5} & \underline{84.1} \\
CoP & 61.0 & 63.1 & 74.0 & 54.3 & 58.8 & 80.3 & 78.2 & 74.8 & 68.1 \\
CoRP & 63.4 & 75.9 & 80.6 & 70.8 & 72.6 & 86.2 & 84.3 & 83.6 & 77.2 \\
ReAct & \textbf{72.7} & 80.2 & 83.3 & 77.8 & 81.6 & 91.9 & 81.1 & 86.1 & 81.8 \\
ASH & 49.2 & 38.8 & 28.9 & 39.7 & 41.2 & 24.5 & 42.0 & 30.4 & 36.8 \\
SCALE & 68.6 & 69.9 & 63.6 & 70.5 & 76.6 & 76.4 & 73.1 & 69.3 & 71.0 \\
ViM & 62.5 & 74.5 & 84.5 & 68.0 & 70.4 & 91.8 & 79.5 & 87.9 & 77.4 \\
SHE & 69.4 & 76.5 & 77.0 & \textbf{82.3} & \textbf{84.8} & 79.5 & 84.5 & 75.0 & 78.6 \\
GradNorm & 60.3 & 51.8 & 45.4 & 52.4 & 56.6 & 47.0 & 55.8 & 45.0 & 51.8 \\
NNGuide & 65.6 & 77.7 & 82.6 & 78.8 & 81.0 & 85.6 & 85.0 & 83.8 & 80.0 \\
ComboOOD & 52.5 & 62.9 & \textbf{87.4} & 59.0 & 63.8 & 80.0 & \textbf{88.3} & 75.7 & 71.2 \\
\midrule
\textbf{\mine} & 70.1 & \underline{83.5} & \underline{87.1} & \underline{81.8} & \underline{84.4} & \underline{92.4} & 86.3 & \underline{88.9} & \textbf{84.3} \\
\bottomrule
\end{tabular}}
\end{table}

\begin{table}[t]
\centering
\caption{Per-OOD-dataset FPR@95\,(\%)$\downarrow$ --- ViT-B / ImageNet.}
\label{tab:full-vit-fpr}
{\footnotesize\setlength{\tabcolsep}{2.5pt}
\begin{tabular}{@{}l ccc ccccc c@{}}
\toprule
& \multicolumn{3}{c}{Near-OOD} & \multicolumn{5}{c}{Far-OOD} & \\
\cmidrule(lr){2-4}\cmidrule(lr){5-9}
Method & SSB & NINCO & ImgO & Places & SUN & iNat & Text & OImgO & Avg. \\
\midrule
MSP & 84.6 & 72.0 & 69.3 & 71.2 & 67.6 & 49.9 & 70.2 & 60.6 & 68.2 \\
Energy & 75.4 & 66.7 & 59.2 & \underline{67.4} & \textbf{59.4} & 47.4 & 57.3 & 53.1 & \underline{60.7} \\
Max Logit & 78.9 & 67.0 & 59.4 & \textbf{65.9} & \underline{60.0} & 43.0 & 61.4 & 52.8 & 61.0 \\
KNN & 98.3 & 98.3 & 98.5 & 99.0 & 99.0 & 98.0 & 92.3 & 98.0 & 97.7 \\
Mahalanobis & 92.4 & 86.9 & 84.7 & 90.5 & 91.9 & 71.1 & 85.6 & 69.5 & 84.1 \\
MDS++ & \underline{75.3} & \underline{66.6} & 70.5 & 82.6 & 77.8 & \textbf{28.3} & 60.4 & \underline{47.5} & 63.6 \\
CoP & 91.1 & 91.8 & 76.3 & 95.5 & 94.2 & 78.4 & 77.1 & 81.9 & 85.8 \\
CoRP & 89.5 & 84.5 & 74.0 & 90.0 & 89.5 & 73.4 & 67.1 & 72.6 & 80.1 \\
ReAct & \textbf{74.8} & \textbf{65.0} & \underline{53.1} & 67.7 & 61.2 & \underline{36.0} & 56.5 & \textbf{47.1} & \textbf{57.7} \\
ASH & 95.6 & 98.6 & 99.5 & 98.9 & 98.8 & 99.8 & 97.4 & 99.4 & 98.5 \\
SCALE & 76.5 & 75.4 & 80.4 & 77.2 & 68.7 & 67.8 & 67.7 & 68.6 & 72.8 \\
ViM & 90.8 & 85.0 & 77.7 & 88.6 & 88.8 & 58.1 & 80.7 & 62.8 & 79.1 \\
SHE & 80.8 & 73.4 & 66.6 & 67.9 & 62.6 & 59.4 & \underline{51.9} & 65.0 & 66.0 \\
GradNorm & 85.9 & 91.1 & 92.7 & 92.8 & 90.9 & 93.5 & 85.9 & 89.3 & 90.3 \\
NNGuide & 88.3 & 82.0 & 75.2 & 77.8 & 75.6 & 75.1 & 57.4 & 66.7 & 74.8 \\
ComboOOD & 90.7 & 81.1 & \textbf{50.3} & 85.6 & 82.5 & 67.6 & \textbf{43.0} & 69.3 & 71.3 \\
\midrule
\textbf{\mine} & 81.6 & 69.8 & 57.5 & 69.4 & 64.8 & 44.3 & 54.3 & 53.1 & 61.8 \\
\bottomrule
\end{tabular}}
\end{table}

\begin{table}[t]
\centering
\caption{Per-OOD-dataset AUROC~(\%)$\uparrow$ --- Swin-B / ImageNet.}
\label{tab:full-swin-auroc}
{\footnotesize\setlength{\tabcolsep}{2.5pt}
\begin{tabular}{@{}l ccc ccccc c@{}}
\toprule
& \multicolumn{3}{c}{Near-OOD} & \multicolumn{5}{c}{Far-OOD} & \\
\cmidrule(lr){2-4}\cmidrule(lr){5-9}
Method & SSB & NINCO & ImgO & Places & SUN & iNat & Text & OImgO & Avg. \\
\midrule
MSP & 76.4 & 85.8 & 80.6 & 82.7 & 84.3 & 94.8 & 83.8 & 91.3 & 85.0 \\
Energy & 79.1 & 84.4 & 82.8 & 77.0 & 81.3 & 95.2 & 81.8 & 90.7 & 84.0 \\
Max Logit & 78.8 & 85.9 & 82.5 & 80.1 & 83.4 & 95.7 & 83.5 & 91.8 & 85.2 \\
KNN & 72.1 & 76.8 & 84.9 & 81.2 & 84.5 & 91.5 & 90.0 & 88.1 & 83.6 \\
Mahalanobis & 75.0 & 86.9 & 90.9 & 84.1 & 85.4 & 98.6 & 88.2 & 94.5 & 87.9 \\
MDS++ & 79.9 & \textbf{90.9} & 93.3 & 80.8 & 82.9 & 98.8 & 91.2 & \underline{96.3} & 89.3 \\
CoP & \underline{82.4} & 88.0 & \underline{95.4} & 74.6 & 78.9 & \underline{99.0} & \underline{92.3} & 95.0 & 88.2 \\
CoRP & 77.7 & 86.2 & 91.8 & 78.2 & 80.5 & 95.7 & 89.8 & 93.5 & 86.7 \\
ReAct & 79.9 & 86.3 & 86.3 & 80.7 & 84.6 & 97.5 & 84.5 & 93.4 & 86.7 \\
ASH & 62.1 & 50.4 & 55.1 & 43.4 & 52.1 & 77.6 & 28.3 & 58.0 & 53.4 \\
SCALE & 79.0 & 85.2 & 83.9 & 79.1 & 82.7 & 95.9 & 82.2 & 92.1 & 85.0 \\
ViM & 79.3 & 88.1 & 93.4 & \underline{84.3} & \underline{86.6} & \textbf{99.1} & 90.2 & 95.7 & \underline{89.6} \\
SHE & 74.4 & 82.6 & 87.9 & 59.2 & 63.1 & 92.5 & 90.1 & 89.7 & 79.9 \\
GradNorm & 54.5 & 45.9 & 33.7 & 30.9 & 34.5 & 33.7 & 38.0 & 43.2 & 39.3 \\
NNGuide & 77.3 & 88.1 & 93.0 & 82.4 & 84.7 & 98.1 & 91.0 & 96.0 & 88.8 \\
ComboOOD & 70.7 & 74.6 & \textbf{96.5} & 67.5 & 75.3 & 90.3 & \textbf{92.4} & 86.4 & 81.7 \\
\midrule
\textbf{\mine} & \textbf{83.1} & \underline{90.9} & 92.6 & \textbf{85.0} & \textbf{87.9} & 98.6 & 91.4 & \textbf{96.4} & \textbf{90.7} \\
\bottomrule
\end{tabular}}
\end{table}

\begin{table}[t]
\centering
\caption{Per-OOD-dataset FPR@95\,(\%)$\downarrow$ --- Swin-B / ImageNet.}
\label{tab:full-swin-fpr}
{\footnotesize\setlength{\tabcolsep}{2.5pt}
\begin{tabular}{@{}l ccc ccccc c@{}}
\toprule
& \multicolumn{3}{c}{Near-OOD} & \multicolumn{5}{c}{Far-OOD} & \\
\cmidrule(lr){2-4}\cmidrule(lr){5-9}
Method & SSB & NINCO & ImgO & Places & SUN & iNat & Text & OImgO & Avg. \\
\midrule
MSP & 70.2 & 51.9 & 59.0 & 61.5 & 57.9 & 22.9 & 54.1 & 35.1 & 51.6 \\
Energy & \underline{58.6} & 44.8 & 46.4 & 59.6 & 51.3 & 15.5 & 53.8 & 28.3 & 44.8 \\
Max Logit & 62.6 & 45.1 & 49.6 & 57.2 & 51.5 & 15.4 & 49.9 & 27.4 & 44.8 \\
KNN & 78.7 & 82.5 & 72.1 & 81.1 & 69.6 & 58.3 & 44.0 & 62.3 & 68.6 \\
Mahalanobis & 81.4 & 68.1 & 52.8 & 69.0 & 68.5 & 6.2 & 51.1 & 33.0 & 53.8 \\
MDS++ & 60.8 & 44.2 & 36.4 & 80.3 & 77.5 & 5.8 & 33.8 & 21.5 & 45.0 \\
CoP & \textbf{58.4} & 53.0 & \underline{26.8} & 84.6 & 78.3 & \underline{5.2} & \textbf{32.2} & 27.8 & 45.8 \\
CoRP & 73.4 & 67.7 & 51.1 & 85.5 & 82.8 & 30.2 & 41.6 & 41.5 & 59.2 \\
ReAct & 59.0 & \underline{43.8} & 41.5 & \underline{55.6} & \textbf{47.6} & 10.0 & 50.0 & 23.5 & \underline{41.4} \\
ASH & 82.3 & 82.7 & 80.3 & 88.2 & 81.6 & 44.7 & 97.0 & 63.6 & 77.5 \\
SCALE & 60.4 & 45.5 & 46.5 & 58.6 & 51.5 & 14.6 & 54.2 & 26.9 & 44.8 \\
ViM & 75.2 & 61.8 & 40.5 & 65.9 & 63.2 & \textbf{3.1} & 42.9 & 25.0 & 47.2 \\
SHE & 68.4 & 55.0 & 47.0 & 75.1 & 72.4 & 24.6 & 36.2 & 35.2 & 51.7 \\
GradNorm & 80.4 & 81.9 & 91.4 & 92.6 & 89.1 & 88.8 & 92.6 & 79.5 & 87.1 \\
NNGuide & 70.7 & 53.4 & 36.4 & 70.2 & 68.3 & 8.5 & 33.6 & \underline{20.2} & 45.1 \\
ComboOOD & 79.5 & 72.5 & \textbf{18.6} & 81.2 & 72.7 & 43.7 & \underline{32.6} & 52.6 & 56.7 \\
\midrule
\textbf{\mine} & 58.7 & \textbf{40.8} & 35.8 & \textbf{54.9} & \underline{47.9} & 6.4 & 35.7 & \textbf{17.7} & \textbf{37.2} \\
\bottomrule
\end{tabular}}
\end{table}

\end{document}